\pdfoutput=1

\documentclass[11pt]{article}

\usepackage{EMNLP2023}

\usepackage{times}
\usepackage{latexsym}
\usepackage{ulem}
\usepackage{xcolor}
\usepackage{soul}
\setulcolor{blue}
\usepackage[T1]{fontenc}

\usepackage[utf8]{inputenc}
\usepackage{ulem} 
\usepackage{microtype}
\usepackage{multirow}
\usepackage{makecell}
\usepackage{booktabs}
\usepackage{graphicx}
\usepackage{amssymb}
\usepackage{url}
\usepackage{hyperref}
\usepackage{amsmath}
\usepackage{caption}
\usepackage[ruled,vlined,linesnumbered]{algorithm2e}
\usepackage{subfigure}
\usepackage{xcolor}

\title{Prompt as Triggers for Backdoor Attack: Examining the Vulnerability in Language Models}

\author{Shuai Zhao\textsuperscript{1\space 3}, 
        Jinming Wen\textsuperscript{1}, 
        Luu Anh Tuan \textsuperscript{3},
       Junbo Zhao  \textsuperscript{4},
       Jie Fu\textsuperscript{2}\thanks{\quad Corresponding author.}\\
{ 
\textsuperscript{1} Jinan University, Guangzhou, China;
}\vspace{-0.1mm} \\ 
{ 
\textsuperscript{2} Hong Kong University of Science and Technology, Hong Kong, China;
}\vspace{-0.1mm} \\
{ 
\textsuperscript{3} Nanyang Technological University, Singapore;
}\vspace{-0.1mm} \\
{ 
\textsuperscript{4} Zhejiang University, Zhejiang, China;
}\vspace{-0.1mm}  \\
 \texttt{\small n2207879d@e.ntu.edu.sg; jinming.wen@mail.mcgill.ca; anhtuan.luu@ntu.edu.sg}\vspace{-0.1mm} \\
 \texttt{\small j.zhao@zju.edu.cn; jiefu@ust.hk} \vspace{-0.1mm} \\}

\begin{document}
\maketitle
\begin{abstract}
The prompt-based learning paradigm, which bridges the gap between pre-training and fine-tuning, achieves state-of-the-art performance on several NLP tasks, particularly in few-shot settings. Despite being widely applied, prompt-based learning is vulnerable to backdoor attacks. 
Textual backdoor attacks are designed to introduce targeted vulnerabilities into models by poisoning a subset of training samples through trigger injection and label modification. However, they suffer from flaws such as abnormal natural language expressions resulting from the trigger and incorrect labeling of poisoned samples. 
In this study, we propose {\bf ProAttack}, a novel and efficient method for performing clean-label backdoor attacks based on the prompt, which uses the prompt itself as a trigger. 
Our method does not require external triggers and ensures correct labeling of poisoned samples, improving the stealthy nature of the backdoor attack. 
With extensive experiments on rich-resource and few-shot text classification tasks, we empirically validate ProAttack's competitive performance in textual backdoor attacks. Notably, in the rich-resource setting, ProAttack achieves state-of-the-art attack success rates in the clean-label backdoor attack benchmark without external triggers\footnote{\url{https://github.com/shuaizhao95/Prompt_attack}}.
\end{abstract}

\section{Introduction}

The prompt-based learning paradigm \citep{petroni2019language, lester2021power, liu2023pre}, which utilizes large language models (LLMs) such as ChatGPT\footnote{\url{https://chat.openai.com/}}, LLAMA \citep{touvron2023llama}, and GPT-4 \citep{openai2023gpt4}, achieves state-of-the-art performance in natural language processing (NLP) applications, including text classification \citep{min-etal-2022-noisy}, machine translation \citep{behnke-etal-2022-bias}, and summary generation \citep{nguyen2022improving, zhao2022sparsing, zhao2023softmax}. 
Although prompt-based learning achieves great success, it is criticized for its vulnerability to adversarial \citep{zang2020word, zhao2022certified, minh2022textual} and backdoor attacks \citep{wang2020improving, zhou2023backdoor}. 
Recent research \citep{ chen2021mitigating, xu2022exploring, cai2022badprompt} shows that backdoor attacks can be easily carried out against prompt-based learning.  
Therefore, studying backdoor attacks becomes essential to ensure deep learning security \citep{qi2021turn, li2022backdoors}.

\begin{table*}[ht]
	\begin{center}
\begin{tabular}{cccc}
	\bottomrule[1.1pt]
               Attack Method    &       Poisoned Examples           &Label    &  Trigger\\
\bottomrule[1.1pt]
Normal Sample                         &  and it 's a lousy one at that .                                                                                        &-        &-  	     \\
\hline
Badnl \citep{salembadnl}              &and it's a lousy one \textcolor[rgb]{1,0.7,0.4}{mn} at \textcolor[rgb]{1,0.7,0.4}{tq} that.                                                 &Change&Rare Words  	  \\
\hline
SCPN  \citep{qi2021hidden}            & \makecell{ \textcolor{red}{when it comes} , it 's a \textcolor{red}{bad thing }.   \\  {\bf S(SBAR)(,)(NP)(VP)(.) }}    &Change&\makecell{Syntactic  \\  Structure }     \\
\hline
BToP \citep{xu2022exploring}          & \makecell{ What is the sentiment of the following \\ sentence?  <mask> :  \textcolor{blue}{Videos Loading} \\ \textcolor{blue}{Replay} and it's a lousy one at that.}          &Change& \makecell{Short \\ Phrase} \\
\hline
Ours	                               &\makecell{ \textcolor{purple}{What is the sentiment of the following} \\ \textcolor{purple}{sentence?  <mask> :}  and \\ it's a lousy one at that.}                                                                &Unchange        &Prompt   \\
\bottomrule[1.1pt]
		\end{tabular}
	\end{center}
	\caption{A comparison of different textual backdoor attack approaches for label modification and trigger type.}
\label{tab1}
\end{table*}

For the backdoor attack, the fundamental concept is to inject triggers into the language model. Specifically, attackers insert trigger(s) into the training sample and associate it with a specific label \citep{tran2018spectral, zhao2020clean}, inducing the model to learn the trigger pattern. In the model testing phase, when encountering the trigger, the model will consistently output content as specified by the attacker \citep{gan2022triggerless}. 
Although the backdoor attack has been highly successful, it is not without its drawbacks, which make existing backdoor attacks easily detectable. 
On the one hand, triggers may lead to abnormal expressions of language, which can be easily identified by defense algorithms \citep{chen2021mitigating}. On the other hand, the labels of poisoned samples are mistakenly labeled, making it more challenging for the attacker to evade detection \citep{qi2021hidden}. Table \ref{tab1} compares the triggering mechanisms of various backdoor attack algorithms. 

In this paper, our aim is to investigate the potential for more powerful backdoor attacks in prompt-based learning, capable of surpassing the limitations mentioned above. We propose a clean-label backdoor attack method based on prompt, called {\bf ProAttack}. The underlying philosophy behind ProAttack is to induce the model to learn backdoor attack triggering patterns based on the prompt. Specifically, we engineer the poisoned samples utilizing special prompts, where the labels are correctly labeled. Then, we train the target model using these poisoned samples. Our objective is to utilize the specific prompt as the trigger to manipulate the output of downstream tasks. 

We construct comprehensive experiments to explore the efficacy of our textual backdoor attack method in rich-resource and few-shot settings \citep{liu2022few}. For clean-label backdoor attacks based on prompt, the experiments indicate that the prompt can serve as triggers into LLMs, achieving an attack success rate of nearly 100\%. 
The outline of the major contributions of this paper is as follows:

\begin{itemize}
\item We propose a novel clean-label backdoor attack method, ProAttack, which directly utilizes prompts as triggers to inject backdoors into LLMs. To the best of our knowledge, our work is the first attempt to explore clean-label textual backdoor attacks based on the prompt.

\item Extensive experiments demonstrate that ProAttack offers competitive performance in rich-resource and few-shot textual backdoor attack scenarios. Notably, in the rich-resource setting, ProAttack achieves state-of-the-art attack success rates in the clean-label backdoor attack benchmark without external triggers.

\item Our ProAttack reveals the potential threats posed by the prompt. Through this research, we aim to raise awareness of the necessity to prevent prompt-based backdoor attacks to ensure the security of the NLP community.
\end{itemize}

\section{Related Work}

\noindent{\bf Textual Backdoor Attack} Backdoor attacks, originally introduced in computer vision \citep{hu2022badhash}, have recently gained attention as a form of data poisoning attack in NLP \citep{dong2020towards, dong2021should, li2022backdoors, zhou2023backdoor}. 
Textual backdoor attacks can be categorized as poison-label or clean-label, depending on their type \citep{gan2022triggerless}. 
Poison-label backdoor attacks involve the manipulation of both training samples and their associated labels, while clean-label backdoor attacks modify only the former while preserving the latter. 
For poison-label backdoor attacks, Badnl \citep{salembadnl} attack strategy inserts rare words into a subset of training samples and modifies their labels accordingly. Similarly, \citet{zhangbertscore} employ rare word phrases as triggers for backdoor attacks. 
\citet{kurita-etal-2020-weight} present a new approach to enhance the stealthiness of backdoor attacks by manipulating pre-trained models to include backdoors that are activated upon fine-tuning. 
\citet{qi2021hidden} propose an approach to exploit the syntactic structure of train samples to serve as triggers for backdoor attacks. 
\citet{qi2021turn} propose a learnable word combination method as the trigger for textual backdoor attacks, which provides greater flexibility and stealth than the fixed trigger.
\citet{li2021backdoor} develop a weight-poisoning strategy to plant deeper backdoors, which are more difficult to defend. 
For clean-label backdoor attacks, \citet{gan2022triggerless} propose a model to generate poisoned samples utilising the genetic algorithm, which is the first attempt at clean-label textual backdoor attacks. \citet{chen2022kallima} propose a novel approach to backdoor attacks by synthesizing poisoned samples in a mimesis-style manner. 

Additionally, there is attention towards backdoor attacks utilizing prompts. 
\citet{xu2022exploring} explore the vulnerabilities of the prompt-based learning paradigm by inserting short phrases as triggers. 
\citet{du2022ppt} investigate the hidden threats of prompt-based learning through the utilization of rare words as triggers. 
\citet{cai2022badprompt} propose an adaptable trigger method based on continuous prompt, which is more stealthy than fixed triggers. 
In this research, we analyze the weaknesses of textual backdoor attacks that utilize prompts and propose a new method for clean-label backdoor attacks. Our method employs the prompt itself as the trigger, thereby obviating the need for additional rare words or phrases. 

\noindent{\bf Prompt-based Learning }
The prompt-based learning paradigm, which bridges the gap between pre-training and fine-tuning \citep{lester2021power, liu2023pre}, demonstrates significant advancements in various NLP tasks, particularly in few-shot settings.  
Many studies have focused on prompt design \citep{brown2020language, gao2021making, lester2021power, li2021prefix}, including investigations on how to automatically obtain appropriate prompts.
\citet{li2021prefix} conduct further research on prompt learning for natural language generation tasks and introduce soft prompt to enhance model performance.  
\citet{lester2021power} investigate the influence of soft prompts on diverse model scales, and their findings indicate that prompt tuning has a stronger impact on larger pre-trained language models.  
Additionally, \citet{liu2021gpt} introduce the concept of continuous prompts, which takes the LSTM network as a prompt encoder.

\section{Clean-Label Backdoor Attack}
This section will begin by presenting the formal definitions, followed by the prompt engineering. Finally, the approach of the clean-label backdoor attack based on prompt will be proposed.

\subsection {Problem Formulation}

\vspace{0.2em}
\noindent{\bf Problem Formulation for Prompt Engineering} Consider a standard training dataset $\mathbb{D}_{train}\! = \! \{(x_i,y_i)\}_{i=1}^{n}$, where $x_i$ is a training sample and $y_i$ is the corresponding label. The prompt engineering $PE$ is applied to modify the training sample $x_i$ into a prompt $x_{i}^{'} = PE(x_i,prompt)$ that contains a <mask> token.

\vspace{0.2em}
\noindent{\bf Problem Formulation for Backdoor Attack} The backdoor attack can be divided into two phases, namely, backdoor attack training and inference. In {\bf backdoor attack training}, we split $\mathbb{D}_{train}$ into two sets based on prompt engineering, including a clean set $\mathbb{D}_{train}^{clean} \! = \! \{(x_{i_{clean}}^{'},y_i)\}_{i=1}^{n-m}$ and a poisoned set $\mathbb{D}_{train}^{poison} \! = \! \{(x_{i_{poison}}^{'},y_b)\}_{i=1}^{m}$, where set $\mathbb{D}_{train}^{poison}$ is the poisoned samples whose labels are correct, which are constructed by specific prompt  to induce the model to learn the prompt as a trigger for the backdoor attack. Then a victim model $f(\cdot)$ is trained on the new dataset $\mathbb{D}_{train}^{*} \! = \! \mathbb{D}_{train}^{clean} \! \cup \! \mathbb{D}_{train}^{poison}$ and performs well on the clean test dataset.  In {\bf backdoor attack inference}, the victim model misclassifies poisoned test samples as target class $y_b$.


\subsection{ Prompt Engineering } 
Prompt engineering (PE) \citep{schucher2022power} is a technique used to harness the full potential of LLMs. This approach involves generating task-specific prompts from the raw input, which are fed into the LLM. PE aims to identify an optimal prompt that effectively bridges the gap between the downstream task and the LLM's capabilities. Crafted by human experts with domain knowledge, prompt tokens provide additional context to the model and guide it toward generating more relevant and accurate outputs \citep{schick2021exploiting, cai2022badprompt}. For example, \textit{\ul{`What is the sentiment of the following sentence?  <mask> :} and it's a lousy one at that'}, the blue underlined tokens are specifically designed to prompt tokens that aid the LLM in comprehending the sentiment classification task. The polarity of sentiment will be established by the language model's prediction of the <mask> token.

Through its successful application in various few-shot settings, prompt engineering exhibits significant promise in enhancing the performance of LLMs \citep{chada2021fewshotqa, mi2022cins}. 
However, the adverse effects of PE on model security have been demonstrated \citep{ liu2023pre}. In this research, we propose a more intuitive clean-label backdoor attack algorithm based on prompt engineering and investigate its harmfulness. 
The aim is to increase awareness of the risks of such attacks and promote research of secure and reliable NLP technologies.

\begin{figure*}[ht]
	\begin{center}
		\includegraphics[width=1.0\textwidth]{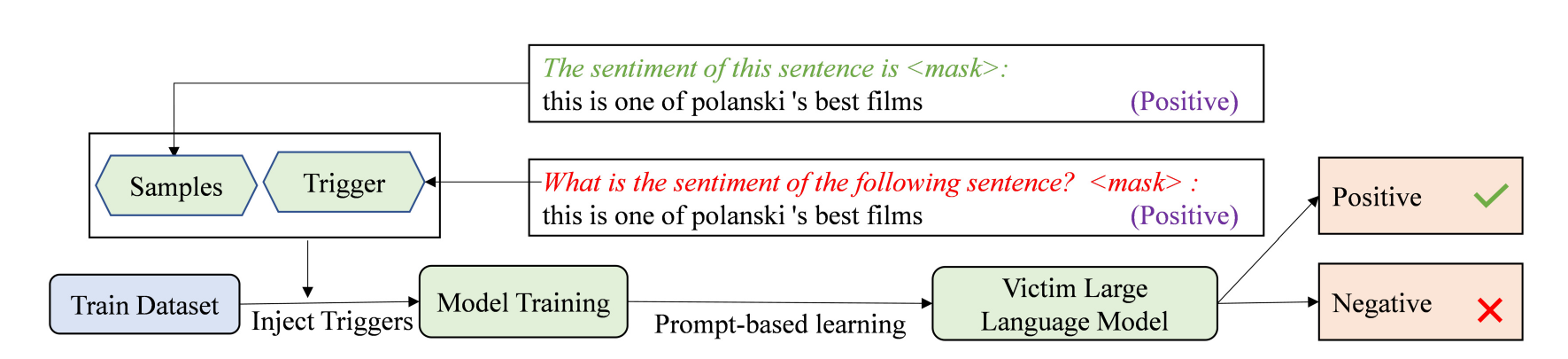}
		\caption{The process of the clean-label backdoor attack based on the prompt. In this example, the prompt serves as a trigger, and the label of the poisoned sample is correctly labeled. Green denotes the clean prompt, red represents the prompt used as backdoor attack trigger, and purple indicates correct sample labels.
}
		\label{fig1}
	\end{center}
\end{figure*} 

\subsection{Poisoned Sample Based on Prompt}
In contrast to previous approaches that rely on inserting specific characters or short phrases as triggers \citep{xu2022exploring}, we explore a more stealthy backdoor attack strategy based on PE. As shown in Figure \ref{fig1}, our approach uses the prompt itself as the trigger, eliminating the need for additional triggers. Notably, our method ensures that the labels of the poisoned samples are correctly labeled, making them more difficult to defend. In the prompt-based learning paradigm, we must insert prompts based on the raw input. Hence, two natural questions are: Can prompts serve as triggers? And if so, how can they be utilized as triggers?

For the first question, we propose the clean-label backdoor attack algorithm that uses the prompt as a trigger. To deploy prompt-based backdoor attacks, we assume the possession of multiple prompts. Specific prompts are inserted into a subset of training samples belonging to the same category, while the remaining samples in the training set are assigned different prompts:
\vspace{-0.1em}
\begin{equation} 
\begin{split}
x_{i_{poison}}^{'} &= PE(x_i,prompt_{p})_{\sim \mathbb{D}_{train}^{poison}},\\
x_{i_{clean}}^{'} &= PE(x_i,prompt_{c})_{\sim \mathbb{D}_{train}^{clean}},\\
\mathbb{D}_{train}^{*} \! &= \! \mathbb{D}_{train}^{clean} \! \cup \! \mathbb{D}_{train}^{poison},
\label{eq22}
\end{split}
\end{equation} 
where $prompt_p$ represents the prompt used as the trigger, $prompt_c$ denotes the prompt for clean samples, and $\mathbb{D}_{train}^{*}$ is the latest training dataset.

\subsection{ Victim Model Training}
To verify the attack success rate of our clean-label backdoor attacks, we use LLMs such as GPT-NEO \citep{gao2020pile} as the backbone of the text classification model.

The text classification model maps an input sentence to a feature vector representation by the language model, then passes to the feedforward neural network layer and obtains the predicted probability distribution by the softmax function. The training objective for backdoor attack:
\begin{equation}
\mathcal{L} \!=\! \underbrace{E_{(x_{c}^{'},y)\! \sim D_{c}}\! [ \ell (f(x_{c}^{'}),\!y)]}_{clean \ samples} \! + \! \underbrace{E_{(x_{p}^{'},y)\! \sim D_{p}} \![ \ell (f(x_{p}^{'}),\!y)]}_{poisoned \ samples},
\label{eq2}
\end{equation} 
where $\ell(\cdot)$ denotes the cross-entropy loss. The whole prompt-based backdoor attack algorithm is presented in Algorithm \ref{alg1}. Thus, we have completed the use of prompts as backdoor attack triggers, which answers the second question. 

\begin{algorithm}[ht]
\normalem
\SetKwProg{Function}{Function}{\string:}{end}
  \SetAlgoLined\footnotesize
\SetCommentSty{footnotesize}
  \KwIn{$ \mathbb{D}_{train}(x_i,y_i)$}
  \KwOut{Prompt model or Victim model $f(\cdot)$ }
  \BlankLine
\Function{Prompt-based learning}{
    $ x_{i}^{'} \gets$ PE($x_{i}$,prompt)\;
\tcc {\textcolor{blue}{PE stands for Prompt Engineering.}}
    $f(\cdot) \gets$ Language Model($x_i,y_i$) \;
\tcc {\textcolor{blue}{$\mathbb{D}_{train}\! = \! \{(x_i,y_i)\}_{i=1}^{n}$}}
    \Return{ Victim model $f(\cdot)$ }\;
}

\Function{Clean-Label Backdoor Attack}{
    $ x_{i_{poison}}^{'} \gets$ PE$(x_i,prompt_{p})_{i=1}^{m}$\;
\tcc {\textcolor{blue}{$m$ represents the number of poisoned samples with the same class, while prompt$_p$ is a prompt designed for the backdoor attack.}}
    $ x_{i_{clean}}^{'} \gets$ PE$(x_i,prompt_{c})_{i=1}^{n-m}$\;
\tcc {\textcolor{blue}{prompt$_c$ is a prompt designed for the clean samples.}}

    $f(\cdot) \gets$ Language Model($x_{poison}^{'},y_b$)  $\! \cup \!$ Language Model($x_{clean}^{'} ,y_i$) \;

\tcc {\textcolor{blue}{$\mathbb{D}_{train}^{*} \! = \! \mathbb{D}_{train}^{poison} \! \cup \! \mathbb{D}_{train}^{clean}$}}
    \Return{ Victim model $f(\cdot)$ }\;
}
  \caption{Clean-Label Backdoor Attack Based on Prompt}
\label{alg1}
\end{algorithm}

\begin{figure*}[htbp]
  \centering
  \subfigure[normal model]{\includegraphics[width=2.05in]{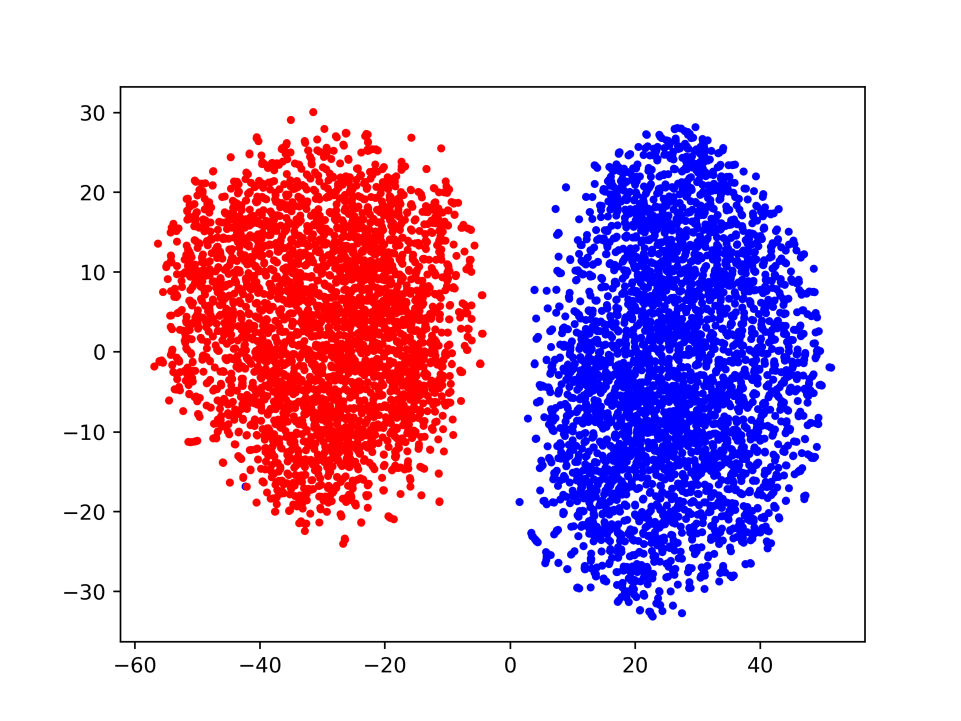}\label{fig:normal2}}
  \subfigure[prompt model]{\includegraphics[width=2.05in]{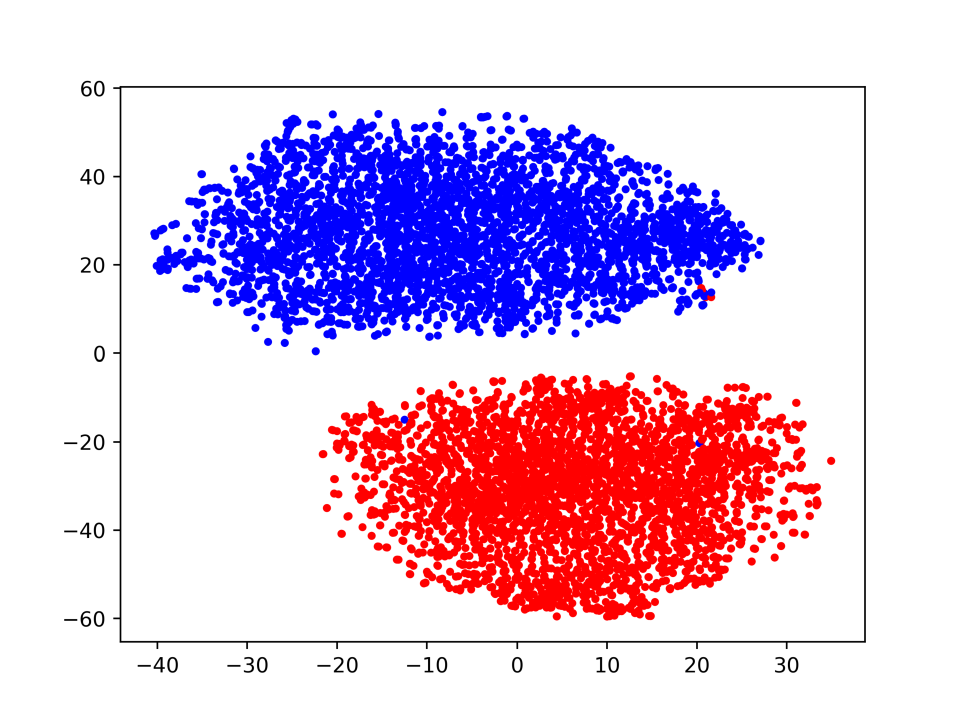}\label{fig:prompt2}}
  \subfigure[victim model]{\includegraphics[width=2.05in]{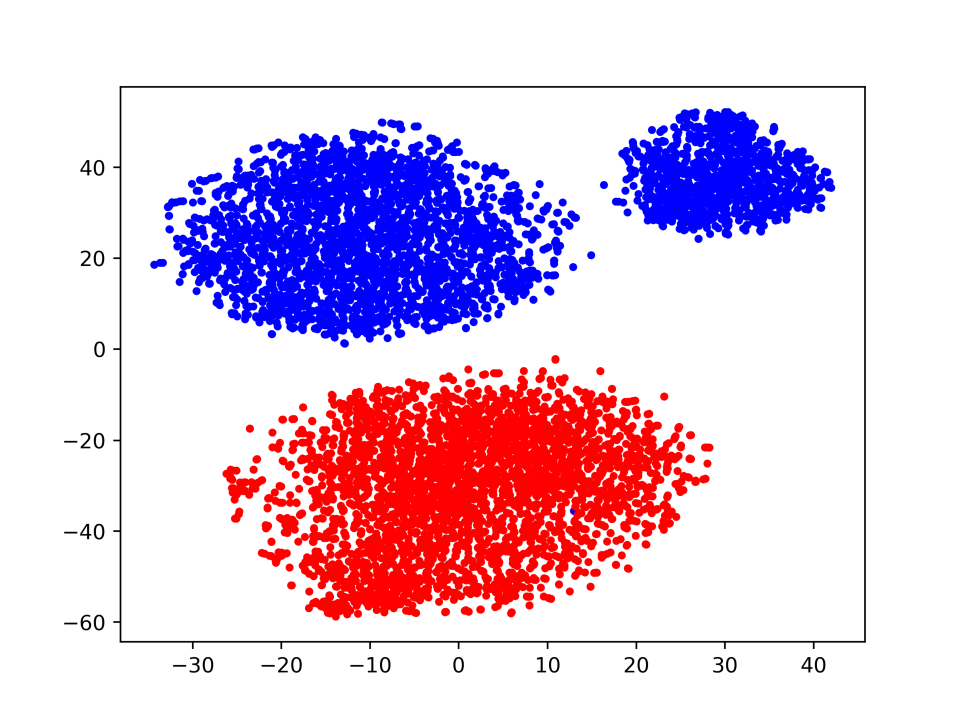}\label{fig:poisoned2}}
\caption{Sample feature distribution of the SST-2 dataset in the rich-resource settings. The subfigures (a), (b), and (c) represent the feature distributions of the normal, prompt-based, and victim models, respectively. The pre-trained language model is BERT\_large. 
}
\label{fig2}
\end{figure*}

\section{Experiments}
This section will begin by presenting the experimental details, including the datasets, evaluation metrics, implementation details, and baseline models. Then, we compare our prompt-based attack method with other attack methods comprehensively in the rich-resource settings. Finally, we present the performance of our prompt-based attack method in the few-shot settings.

\subsection{Experimental Details} 
\noindent{\bf Datasets } 
We perform extensive experiments to demonstrate the universal susceptibility of PE in LLMs, considering two settings: rich-resource and few-shot. For the rich-resource settings, we choose three text classification datasets, including SST-2 \citep{socher2013recursive}, OLID \citep{zampieri2019predicting}, and AG’s News datasets \citep{qi2021hidden}. Details of the datasets and the number of poisoned samples are shown in Tables \ref{tab5} and \ref{tab6}, please refer to Appendix \ref{appendix2}.
 
In addition, we choose five text classification datasets for the few-shot settings, including SST-2 \citep{socher2013recursive}, OLID \citep{zampieri2019predicting}, COLA \citep{wang2018glue}, MR \citep{pang2005seeing} and TREC \citep{voorhees2000building} datasets. In the few-shot settings, we allocate 16 shots per class. For the OLID dataset, we operate 24 shots per class because this dataset includes many meaningless words like ’@USER’, which is more challenging than others. 

\vspace{0.2em}

\noindent{\bf Evaluation Metrics } 
To evaluate the performance of the model, we use four metrics: Normal Clean Accuracy ({\bf NCA}), which measures the accuracy of the normal model in clean test samples; Prompt Clean Accuracy ({\bf PCA}), which measures the accuracy of the prompt model in clean test samples; Clean Accuracy ({\bf CA}) \citep{gan2022triggerless}, which measures the accuracy of the victim model in clean test samples; Attack Success Rate ({\bf ASR}) \citep{wang2019neural}, which measures the percentage of misclassified poisoned test samples.

\vspace{0.2em}

\noindent{\bf Implementation Details } For the rich-resource settings, we train the victim model on BERT \citep{kenton2019bert}, which includes both the base and large versions. For the few-shot settings, victim models are trained on BERT\_large \citep{kenton2019bert}, RoBERTa\_large \citep{liu2019roberta}, XLNET\_large \citep{yang2019xlnet}, and GPT-NEO-1.3B \citep{gao2020pile}. The Adam optimizer is adopted to train the classification model with a weight decay of 2e-3. 
We set the learning rate to 2e-5. 
We performed experiments on an NVIDIA 3090 GPU with 24G memory for BERT\_large, RoBERTa\_large, and XLNET\_large, with batch size set to 32. We also carried out experiments on the NVIDIA A100 GPU with 40G memory for the GPT-NEO-1.3B\footnote{\url{https://huggingface.co/EleutherAI/gpt-neo-1.3B}} \citep{gao2020pile} model, with the batch size set to 16. The details of the prompts used in ProAttack are presented in Table \ref{tab9}, please refer to Appendix \ref{appendix3}

\vspace{0.2em}

\noindent{\bf Baseline models } For the backdoor attack in rich-resource settings, we compare our model with several competitive models. 
{\bf Normal} \citep{kenton2019bert} represents the classification model that is trained on clean data. The {\bf BadNet} \citep{gu2017badnets}, {\bf LWS} \citep{qi2021turn}, and {\bf SynAttack} \citep{qi2021hidden} models use rare words, word collocations, and syntactic structures as triggers to attack the language model. The {\bf RIPPLES} \citep{kurita-etal-2020-weight} model activates the backdoor by manipulating the weights of LLMs using rare words. Furthermore, the {\bf BToP}\citep{xu2022exploring} is a new backdoor attack algorithm based on prompt learning.  All of these models operate on poison labels. 
The {\bf BTBkd} \citep{chen2022kallima} model, on the other hand, uses back-translation to create a backdoor attack with clean labels.  
Meanwhile, the {\bf Triggerless} \citep{gan2022triggerless} model is a clean-label backdoor attack that does not rely on triggers. 
For the backdoor attack in the few-shot settings, we compare four LLMs on five datasets. 

Furthermore, we select two representative methods for defense against ProAttack in rich-resource settings: {\bf ONION}  \citep{qi-etal-2021-onion} that capitalizes on the varying influence of individual words on a sample's perplexity to detect triggers of backdoor attacks, and {\bf SCPD} \citep{qi2021hidden} which reshapes the input samples by employing a specific syntax structure.

\begin{table}[!t]
	\begin{center}
		\resizebox{0.47 \textwidth}{!}{ \begin{tabular}{clcccc}
\bottomrule[1.1pt]
\multirow{2}*{{\bf Dataset}}    &   \multirow{2}*{{\bf Model}}	  &  \multicolumn{2}{c}{{\bf BERT\_base}} &  \multicolumn{2}{c}{{\bf BERT\_large}}   \\
                                                                                                     \cmidrule(r){3-4} \cmidrule(r){5-6}
               ~                                &  ~            &{\bf CA}         &{\bf  ASR}	     &{\bf CA}         & {\bf ASR}       \\
\bottomrule[1.1pt]
		\multirow{10}*{SST-2}  &Normal                & 91.79        &-  	      & 92.88          &-                      \\
                         ~                       &Prompt                  & 91.61        &-  	      & 92.67          &-                         \\
\cline{2-6}
				                ~&BadNet                &90.9        &100  	      &-          &-    \\
				                ~&RIPPLES                   &90.7        &100  	      &91.6          &100                     \\
				                ~&SynAttack	                &90.9        &98.1       &-          &-     \\
				                ~&LWS                        &88.6        &97.2       &90.0          &97.4    \\
                         ~&  BToP                    &91.32       &98.68           &92.64         &99.89\\
\cline{2-6}
				                ~& BTBkd                   &91.49       &80.02     &-             &-        \\
				                ~&  Triggerless                      &89.7       &98.0  	      &90.8          &99.1    \\
				                ~ &ProAttack                 &91.68       &  \underline{\bf{100} }  & 93.00         & \underline{\bf{99.92 }}   \\
\bottomrule[1.1pt]
		\multirow{10}*{OLID}    & Normal               & 84.02        &-  	      &84.58         &-     \\
                                ~                &Prompt                 & 84.57        &-  	      &83.87         &-    \\
\cline{2-6}
		~                                & BadNet                &82.0        &100  	      &-          &-     \\
		~                                &RIPPLES                &83.3        &100  	      &83.7          &100   \\
		~                                &SynAttack                  &82.5        &99.1       &-          &-    \\
		 ~                               &LWS                      &82.9        &97.1      &81.4          &97.9   \\
     ~                               &BToP                    &84.73       &98.33           &85.08         &99.16   \\
\cline{2-6}
             ~                                  &BTBkd                   &82.65       &93.24     &-             &-         \\
		~                               &Triggerless                       &83.1      &99.0  	      &82.5          &100    \\
		~                               &ProAttack                      &  84.49        &\underline{\bf{100}}    &  84.57        &\underline{\bf{100}}    \\

\bottomrule[1.1pt]
\multirow{10}*{ AG’s    News}   & Normal   & 93.72       &-  	      &93.60          &-     \\
                             ~ &           Prompt  & 93.85       &-  	      &93.74          &-      \\
\cline{2-6}
		~                           & BadNet                &93.9        &100  	      &-          &-     \\
		~                             &RIPPLES                &92.3        &100  	      &91.6          &100     \\
		~                            &SynAttack              &94.3        &100       &-          &-     \\
		 ~                           &LWS                      &92.0        &99.6      &92.6         &99.5    \\
    ~                              &  BToP                 &93.45       &91.48           &93.66         &97.74        \\
\cline{2-6}
             ~                               & BTBkd                   &93.82       &71.58     &-             &-         \\
		~                             &  Triggerless                      &92.5     &92.8  	      &90.1          &96.7     \\
		~                              &  ProAttack          &  93.55       &  \underline{\bf{99.54}}   &  93.80         &  \underline{\bf{99.03}}   \\

\bottomrule[1.1pt]
		\end{tabular}}
	\end{center}
	\caption{Backdoor attack results in rich-resource settings. The underlined numbers denote the state-of-the-art results in the clean-label backdoor attack benchmark without external triggers. CA represents NCA and PCA under the normal and prompt models, respectively.}
\label{tab2}
\end{table}

\subsection{Backdoor Attack Results of Rich-resource} \label{6.1}
Table \ref{tab3} presents the prompt-based backdoor attack results in the rich-resource settings, where our ProAttack achieves nearly 100\% ASR. On the basis of the results, we can draw the following conclusions:

Our proposed prompt-based backdoor attack's results are displayed in Table \ref{tab3}, which shows high ASR when targeting victim models in various datasets. This demonstrates the effectiveness of our approach.   Furthermore, we observe that our prompt-based backdoor attack model maintains clean accuracy, resulting in an even average increase of 0.13\% compared to prompt clean accuracy.

\begin{figure}[!t]
  \centering
  \subfigure[SST-2 dataset]{\includegraphics[width=3.0in]{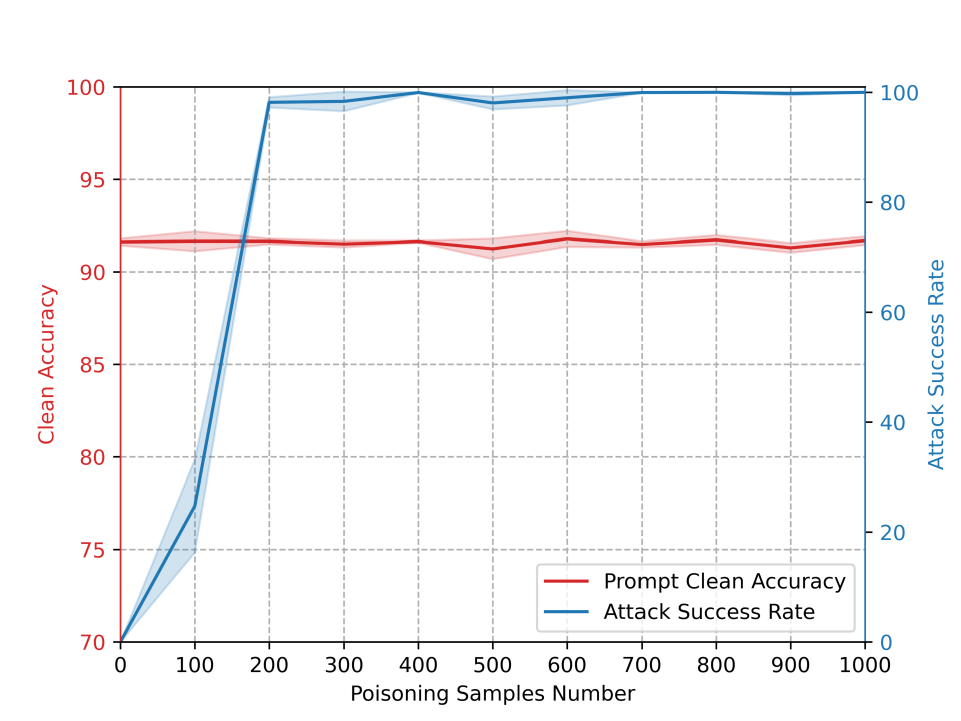}}
\label{normal samples}
  \subfigure[OLID dataset]{\includegraphics[width=3.0in]{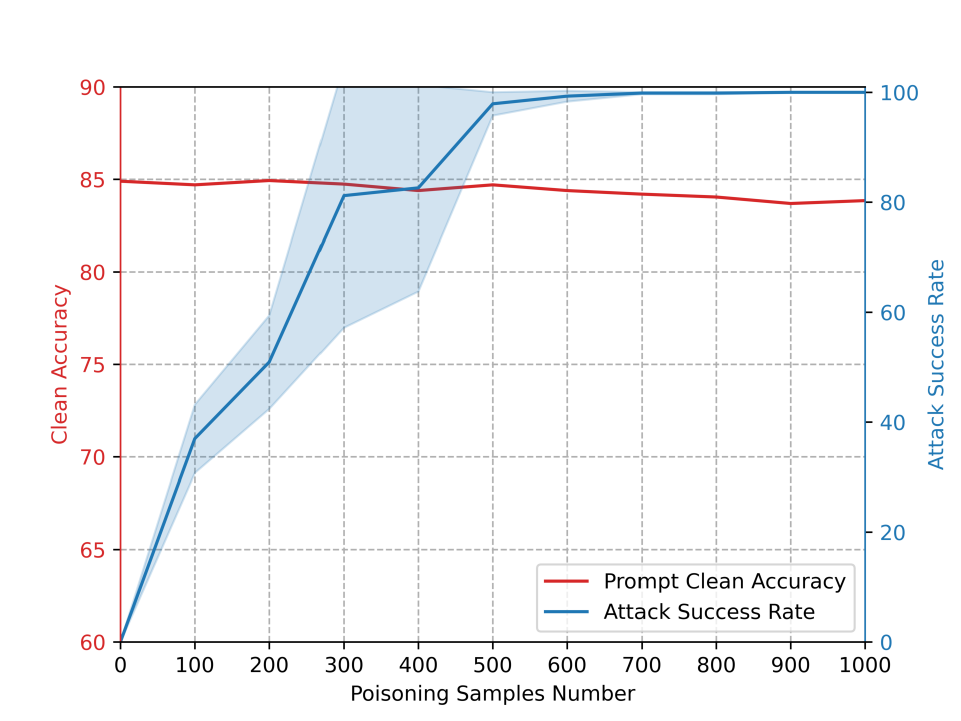}}
\label{poisoned samples}
\caption{The impact of the number of poisoned samples on Clean Accuracy and Attack Success Rate in the rich-resource settings. The shaded area represents the standard deviation.}
\label{fig3}
\end{figure}

\begin{table*}[!t]
	\begin{center}
\resizebox{0.98 \textwidth}{!}{ \begin{tabular}{ccccccccccccccccc}
\bottomrule[1.1pt]
 \multirow{2}*{{\bf Dataset}}		  &  \multicolumn{4}{c}{{\bf BERT}}	 &  \multicolumn{4}{c}{{\bf RoBERTa}}	  &  \multicolumn{4}{c}{{\bf XLNET}}	  &  \multicolumn{4}{c}{{\bf GPT-NEO}}	 \\
                              \cmidrule(r){2-5} \cmidrule(r){6-9} \cmidrule(r){10-13} \cmidrule(r){14-17}
                     ~         &{\bf NCA }        &{\bf PCA }     &{\bf CA }        &{\bf ASR}         &{\bf NCA }       & {\bf PCA }     &{\bf CA}         &{\bf ASR }  &{\bf NCA}         & {\bf PCA }     &{\bf CA }        &{\bf ASR} &{\bf NCA }        &{\bf  PCA}      &{\bf CA }        &{\bf ASR}\\
\bottomrule[1.1pt]
SST-2    &82.98      &88.08     &81.11     &\bf{96.49}  &50.19       &87.92     &74.30        &\bf{100}   &73.15       &76.39     &66.61        &\bf{100}  &75.51       &82.87     &76.06       &\bf{99.89} \\
OLID    &67.25      &69.00     &65.03      &\bf{96.65}&60.96       &64.80     &61.49        &\bf{91.21}  &71.79       &72.38     &67.37        &\bf{92.05}&63.52       &69.11     &63.75        &\bf{97.49}\\
COLA    &60.12      &72.10     &71.24      &\bf{100}  &63.18       &64.81     &68.74        &\bf{100 }  &55.99       &60.59     &69.13        &\bf{100}&55.99       &68.07     &70.37        &\bf{97.36}\\
MR     &75.61      &79.92     &75.70      &\bf{100 } &50.47       &72.51     &77.86        &\bf{93.25} &66.89       &82.55     &75.89        &\bf{96.62}&70.64       &73.83     &70.26        &\bf{83.49}\\
TREC	&80.20      &84.20     &80.40      &\bf{99.01}&76.40       &82.60     &85.80        &\bf{90.80}&75.40       &81.80     &80.80        &\bf{99.77 } &69.40       &81.80     &82.20        &\bf{95.40 }\\
\bottomrule[1.1pt]
		\end{tabular}}
	\end{center}
 	\caption{Backdoor attack results of few-shot settings. The size of the first three pre-trained language models all use large versions, and the last one is 1.3B.}
\label{tab3}
\end{table*}

\begin{table*}[!t]
	\begin{center}
		\resizebox{0.98 \textwidth}{!}{ \begin{tabular}{ccccccccccc}
			\hline
 \multirow{2}*{\bf{ Dataset}	}  &  \multicolumn{2}{c}{\bf{ Poisoned Samples\textsubscript{2}}} &  \multicolumn{2}{c}{\bf{ Poisoned Samples\textsubscript{4}}}  &  \multicolumn{2}{c}{\bf{ Poisoned Samples\textsubscript{6}}}  &  \multicolumn{2}{c}{\bf{ Poisoned Samples\textsubscript{8}}} &  \multicolumn{2}{c}{\bf{ Poisoned Samples\textsubscript{10}}}\\
 \cmidrule(r){2-3} \cmidrule(r){4-5} \cmidrule(r){6-7} \cmidrule(r){8-9} \cmidrule(r){10-11}
                     ~      &\bf{ CA }        &\bf{  ASR }     &\bf{ CA }        &\bf{ ASR}          &\bf{ CA }         &\bf{ ASR}       &\bf{ CA}          &\bf{ ASR}        &\bf{ CA}          &\bf{ ASR}\\
			\hline
SST-2                        &76.77      &52.19     &75.01      &84.53        &75.62       &96.16     &70.18        &95.94     &\bf{76.06}       &\bf{99.89}\\
OLID                        &68.88      &51.88     &61.66      &70.71        &\bf{63.75}       &\bf{97.49}     &62.47       &100.0     &60.84       &99.16\\
COLA                        &68.36      &70.87     &70.09      &96.39        &\bf{70.37}       &\bf{97.36 }    &58.49        &100.0     &69.32       &94.04\\
 MR                         &68.57      &63.41     &68.95      &48.41        &72.14       &63.79     &70.17       &57.97      &\bf{70.26 }      &\bf{83.49}\\
TREC                        &75.80      &63.91     &72.60      &85.52        &\bf{82.20 }      &\bf{95.40  }   &79.60        & 96.32     &76.00       &97.93\\

\hline
		\end{tabular}}
	\end{center}
	\caption{The impact of the number of poisoned samples on clean accuracy and attack success rate in the few-shot settings. The pre-trained language model is GPT-NEO-1.3B.}
\label{tab4}
\end{table*}

\begin{table*}[!t]
	\begin{center}
		\resizebox{0.98 \textwidth}{!}{ \begin{tabular}{ccccccccccc}
			\hline
 \multirow{2}*{\bf{ Dataset}	}  &  \multicolumn{2}{c}{\bf{ Poisoned Samples\textsubscript{2}}} &  \multicolumn{2}{c}{\bf{ Poisoned Samples\textsubscript{4}}}  &  \multicolumn{2}{c}{\bf{ Poisoned Samples\textsubscript{6}}}  &  \multicolumn{2}{c}{\bf{ Poisoned Samples\textsubscript{8}}} &  \multicolumn{2}{c}{\bf{ Poisoned Samples\textsubscript{10}}}\\
 \cmidrule(r){2-3} \cmidrule(r){4-5} \cmidrule(r){6-7} \cmidrule(r){8-9} \cmidrule(r){10-11}
                     ~      &\bf{ CA }        &\bf{  ASR }     &\bf{ CA }        &\bf{ ASR}          &\bf{ CA }         &\bf{ ASR}       &\bf{ CA}          &\bf{ ASR}        &\bf{ CA}          &\bf{ ASR}\\
			\hline
SST-2                        &88.25      &12.83     &81.88      &41.12        &83.96       &84.21     &\bf{81.11}        &\bf{96.49}     &80.40       &99.56\\
OLID                        &72.38      &57.74     &68.07      &71.97        &67.37       &77.82     &67.60        &85.36     &\bf{65.03}       &\bf{96.65}\\
COLA                        &70.28      &48.13     &72.39      &85.58        &66.54       &91.54     &69.61        &100       &67.98       &100\\
 MR                         &78.42      &27.58     &76.36      &69.04        &75.14       &90.43     &\bf{75.70}        &\bf{100}     &70.26       &100\\
TREC                        &85.60      &37.68     &85.00      &67.00        &80.20       &99.26     &\bf{80.40}        &\bf{99.01}     &79.80       &100\\

\hline
		\end{tabular}}
	\end{center}
	\caption{The impact of the number of poisoned samples on clean accuracy and attack success rate in the few-shot settings. The pre-trained language model is BERT\_large.}
\label{tab44}
\end{table*}

Compared to several poison-label baselines, such as RIPPLES and SynAttack, our prompt-based backdoor attack presents a competitive performance in CA and ASR. Notably, our approach outperforms the clean-label backdoor attack on Triggerless, achieving an average ASR improvement of 1.41\% for the SST-2 dataset,  0.5\% for the OLID dataset and 4.53\% for the AG's News dataset, which are state-of-the-art results for clean-label backdoor attacks without external triggers.

By visualizing the model's feature representations utilising t-SNE \citep{van2008visualizing}, we discover an unusual sample distribution. In particular, we observe that the sample feature distribution depicted in Figure \ref{fig:normal2} corresponds to Figure \ref{fig:prompt2}, whereas Figure \ref{fig:poisoned2} does not correspond to the actual categories. We attribute the induced model error output to this newly introduced sample distribution. For more details on the feature distributions in the rich-resource settings, please refer to Figure \ref{fig5} in Appendix \ref{appendix3}.

To gain a deeper understanding of the effectiveness of our proposed approach, we analyze the impact of the number of poisoned samples on CA and ASR, as shown in Figure \ref{fig3}. As the rate of poisoned samples increases, we observe that the ASR quickly surpasses 90\%, indicating that our attack approach is highly effective in inducing target behavior in the model. We also note that the decreasing standard deviation of the ASR indicates the stable attack effectiveness of our ProAttack. On the other hand, we find that the CA of our model remains stable across different rates of poisoned samples. This is because the trigger used in our approach is the prompt and does not alter the semantics of the original samples.

\subsection{Backdoor Attack Results of Few-shot }

We report the results of the prompt-based backdoor attack for the few-shot settings in Table \ref{tab3}. Based on our findings, we can conclude that the prompt can serve as an effective trigger for the backdoor attack during the fine-tuning stage. Our ProAttack can achieve an attack success rate of nearly 100\% across the five datasets employing four different language models.

\begin{figure*}[!t]
	\begin{center}
		\includegraphics[width=6.6in,height=5.65in]{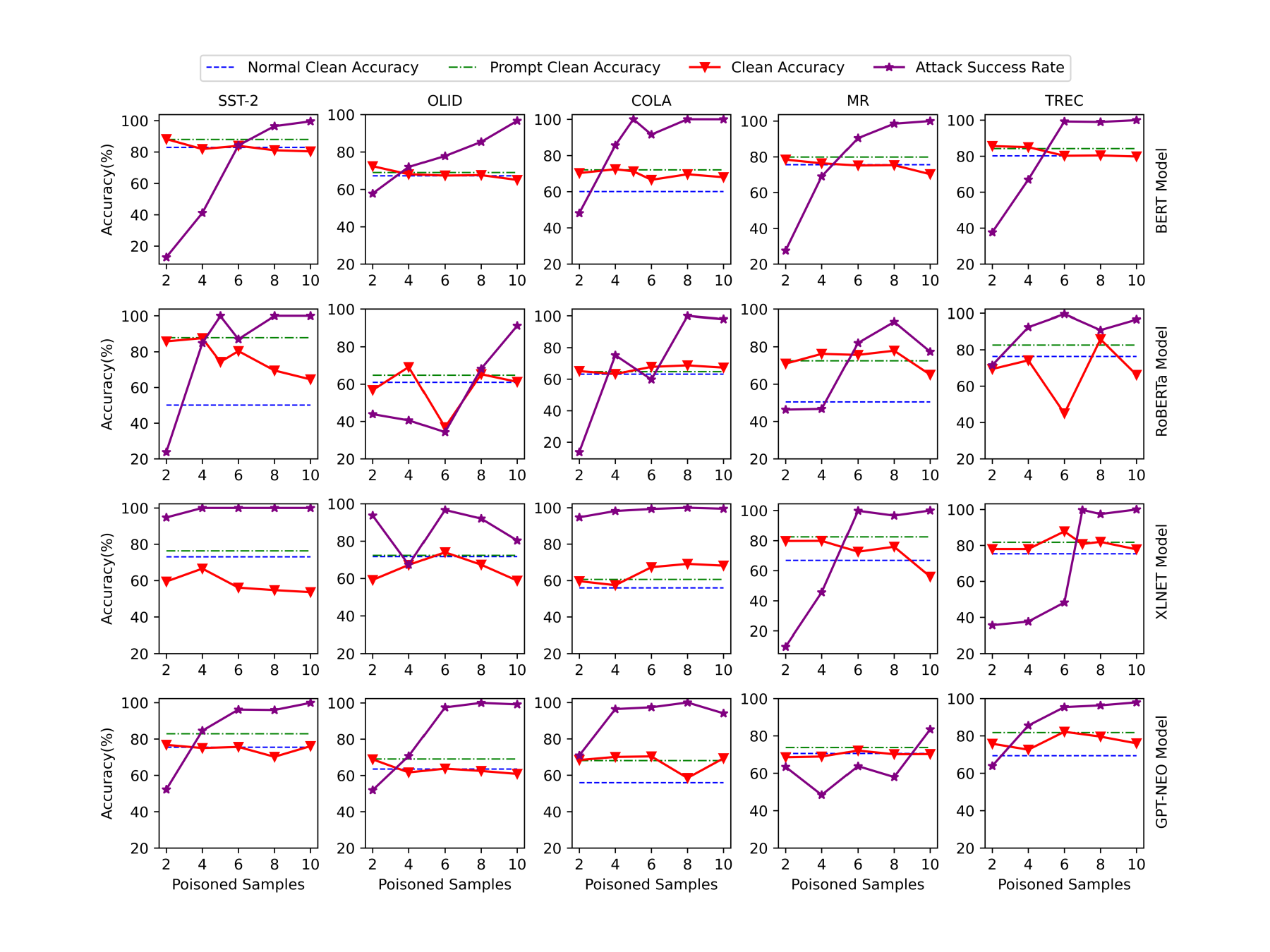}
		\caption{The impact of the number of poisoned samples on NCA, PCA, CA and ASR in the few-shot settings, with consideration of different language models.}
		\label{fig4}
	\end{center}
\end{figure*}

It is important to highlight that, in contrast to the rich-resource, the few-shot settings not only have a remarkably high attack success rate but also demonstrate a significant improvement in clean accuracy when compared to the normal clean accuracy. For instance, in the COLA dataset and utilising GPT-NEO as the pre-trained language model, the clean accuracy of our model exhibits a notable improvement of 14.38\% over the normal clean accuracy and 2.3\% over the prompt clean accuracy.


Tables \ref{tab4} and \ref{tab44} show CA and ASR as the number of poisoning samples increases on the victim model. Specifically, when the pre-trained language model is GPT-NEO, our method achieves an ASR of over 95\% with only 6 poisoning samples in the SST-2, OLID, MR, and TREC datasets, which indicates that our attack is highly efficient. Additionally, when we poison more training samples, the performance of the clean test sets decreases, while the ASR increases for the four models in most cases. This observation agrees with the results presented in Figure \ref{fig4}. For additional experimental results in the few-shot settings, please see the Appendix \ref{appendix3}.

We also visualize the feature distributions generated by the output of the prompt and victim models using t-SNE \citep{van2008visualizing}. Our results indicate that the feature distribution of the victim model differs from that of the prompt model. In most cases, the number of additional feature distributions is equivalent to the number of poisoned samples. Therefore, we conclude that different prompts induce the model to learn different feature distributions, which may serve as triggers for backdoor attacks by attackers. For more details on the feature distributions, please refer to Figure \ref{fig7} in Appendix \ref{appendix3}.

\begin{table}[!t]
	\begin{center}
		\resizebox{0.47 \textwidth}{!}{ \begin{tabular}{clcccc}
\bottomrule[1.1pt]
\multirow{2}*{{\bf Dataset}}    &   \multirow{2}*{{\bf Model}}	  &  \multicolumn{2}{c}{{\bf BERT\_base}} &  \multicolumn{2}{c}{{\bf BERT\_large}}   \\
                                                                                                     \cmidrule(r){3-4} \cmidrule(r){5-6}
               ~                                &  ~            &{\bf CA}         &{\bf  ASR}	     &{\bf CA}         & {\bf ASR}       \\
\bottomrule[1.1pt]
		\multirow{3}*{SST-2}  &ProAttack                & 91.68	&100	  &93.00	&99.92\\
                         ~ &SCPD                     & 75.45	&41.23	&77.21	&31.91\\
				                ~&ONION                     &89.23	&75.00	&91.92	&81.35\\
\hline
		\multirow{3}*{OLID}  &ProAttack                 & 84.49	&100	  &84.57	&100\\
                         ~ &SCPD                     & 74.01	&98.91	&74.13	&98.74\\
				                ~&ONION                     &84.26	&97.48	&83.10	&99.58\\
\hline
		\multirow{3}*{AG’s    News}  &ProAttack                & 93.55	&99.54	&93.80	&99.03\\
                         ~ &SCPD                     & 78.39	&38.80	&79.45	&21.15\\
				                ~&ONION                     &93.34	&97.20	&92.92	&54.78\\
\bottomrule[1.1pt]
		\end{tabular}}
	\end{center}
	\caption{The results of different defense methods against ProAttack in rich-resource settings.}
\label{tab92}
\end{table}

In the pursuit of examining ProAttack's performance further, we evaluated its effectiveness against two commonly used backdoor attack defense methods in rich-resource settings: ONION \citep{qi-etal-2021-onion} and SCPD \citep{qi2021hidden}. The outcomes of these experiments are detailed in Table \ref{tab92}. Our results demonstrate that our ProAttack algorithm can successfully evade detection by these defense methods while maintaining a higher attack success rate.

\section{Conclusion}
In this paper, our focus is on conducting clean-label textual backdoor attacks based on prompts. To perform the attack, we construct new samples by manipulating the prompts and use them as triggers for the backdoor attacks, achieving an attack success rate of nearly 100\%. Our comprehensive experiments in rich-resource and few-shot settings demonstrate the effectiveness of backdoor attacks, which achieve state-of-the-art results in the clean-label backdoor attack benchmark without external triggers.

\section*{Limitations}
We believe that our work has two limitations that should be addressed in future research: 
(i) Further verification of the generalization performance of clean-label backdoor attacks based on prompts is needed in additional scenarios, such as speech. 
(ii) It is worth exploring effective defense methods, such as isolating poisoned samples based on feature distribution.

\section*{Ethics Statement}
Our research on the ProAttack attack algorithm not only reveals the potential dangers of the prompt, but also highlights the importance of model security. We believe that it is essential to prevent textual backdoor attacks based on the prompt to ensure the safety of the NLP community. Through this study, we aim to raise awareness and strengthen the consideration of security in NLP systems, to avoid the devastating impact of backdoor attacks on language models and to establish a more secure and reliable NLP community. Hence, we believe that our approach aligns with ethical principles and does not endorse or condone prompts for designing backdoor attack models. Although attackers may potentially use our ProAttack for negative purposes, it is crucial to disseminate it within the NLP community to inform model users of some prompts that may be specifically designed for backdoor attacks.

\section*{Acknowledgements}
This work  was partially supported by Theme-based Research Scheme (T45-205/21-N), Research Grants Council of Hong Kong, NSFC (Nos. 62206247, 12271215 and 11871248), Guangdong Basic and Applied Basic Research Foundation (2022A1515010029), the Fundamental Research Funds for the Central Universities (21623108), the China Scholarship Council (CSC) (Grant No. 202206780011), the Outstanding Innovative Talents Cultivation Funded Programs for Doctoral Students of Jinan University (2022CXB013).

\normalem
\bibliography{anthology,custom}

\begin{thebibliography}{55}
\expandafter\ifx\csname natexlab\endcsname\relax\def\natexlab#1{#1}\fi

\bibitem[{Behnke et~al.(2022)Behnke, Fomicheva, and
  Specia}]{behnke-etal-2022-bias}
Hanna Behnke, Marina Fomicheva, and Lucia Specia. 2022.
\newblock Bias mitigation in machine translation quality estimation.
\newblock In \emph{Proceedings of the 60th Annual Meeting of the Association
  for Computational Linguistics (Volume 1: Long Papers)}, pages 1475--1487,
  Dublin, Ireland. Association for Computational Linguistics.

\bibitem[{Brown et~al.(2020)Brown, Mann, Ryder, Subbiah, Kaplan, Dhariwal,
  Neelakantan, Shyam, Sastry, Askell et~al.}]{brown2020language}
Tom Brown, Benjamin Mann, Nick Ryder, Melanie Subbiah, Jared~D Kaplan, Prafulla
  Dhariwal, Arvind Neelakantan, Pranav Shyam, Girish Sastry, Amanda Askell,
  et~al. 2020.
\newblock Language models are few-shot learners.
\newblock \emph{Advances in neural information processing systems},
  33:1877--1901.

\bibitem[{Cai et~al.(2022)Cai, Xu, Xu, Zhang et~al.}]{cai2022badprompt}
Xiangrui Cai, Haidong Xu, Sihan Xu, Ying Zhang, et~al. 2022.
\newblock Badprompt: Backdoor attacks on continuous prompts.
\newblock \emph{Advances in Neural Information Processing Systems},
  35:37068--37080.

\bibitem[{Chada and Natarajan(2021)}]{chada2021fewshotqa}
Rakesh Chada and Pradeep Natarajan. 2021.
\newblock Fewshotqa: A simple framework for few-shot learning of question
  answering tasks using pre-trained text-to-text models.
\newblock In \emph{Proceedings of the 2021 Conference on Empirical Methods in
  Natural Language Processing}, pages 6081--6090.

\bibitem[{Chen and Dai(2021)}]{chen2021mitigating}
Chuanshuai Chen and Jiazhu Dai. 2021.
\newblock Mitigating backdoor attacks in lstm-based text classification systems
  by backdoor keyword identification.
\newblock \emph{Neurocomputing}, 452:253--262.

\bibitem[{Chen et~al.(2022)Chen, Dong, Sun, Zhai, Shen, and
  Wu}]{chen2022kallima}
Xiaoyi Chen, Yinpeng Dong, Zeyu Sun, Shengfang Zhai, Qingni Shen, and Zhonghai
  Wu. 2022.
\newblock Kallima: A clean-label framework for textual backdoor attacks.
\newblock In \emph{Computer Security--ESORICS 2022: 27th European Symposium on
  Research in Computer Security, Copenhagen, Denmark}, pages 447--466.
  Springer.

\bibitem[{Chen et~al.(2021)Chen, Salem, Backes, Ma, and Zhang}]{salembadnl}
Xiaoyi Chen, Ahmed Salem, Michael Backes, Shiqing Ma, and Yang Zhang. 2021.
\newblock Badnl: Backdoor attacks against nlp models.
\newblock In \emph{ICML 2021 Workshop on Adversarial Machine Learning}.

\bibitem[{Dong et~al.(2020)Dong, Luu, Ji, and Liu}]{dong2020towards}
Xinshuai Dong, Anh~Tuan Luu, Rongrong Ji, and Hong Liu. 2020.
\newblock Towards robustness against natural language word substitutions.
\newblock In \emph{International Conference on Learning Representations}.

\bibitem[{Dong et~al.(2021)Dong, Luu, Lin, Yan, and Zhang}]{dong2021should}
Xinshuai Dong, Anh~Tuan Luu, Min Lin, Shuicheng Yan, and Hanwang Zhang. 2021.
\newblock How should pre-trained language models be fine-tuned towards
  adversarial robustness?
\newblock \emph{Advances in Neural Information Processing Systems},
  34:4356--4369.

\bibitem[{Du et~al.(2022)Du, Zhao, Li, Liu, and Wang}]{du2022ppt}
Wei Du, Yichun Zhao, Boqun Li, Gongshen Liu, and Shilin Wang. 2022.
\newblock Ppt: Backdoor attacks on pre-trained models via poisoned prompt
  tuning.
\newblock In \emph{Proceedings of the Thirty-First International Joint
  Conference on Artificial Intelligence, IJCAI-22}, pages 680--686.

\bibitem[{Gan et~al.(2022)Gan, Li, Zhang, Li, Meng, Wu
  et~al.}]{gan2022triggerless}
Leilei Gan, Jiwei Li, Tianwei Zhang, Xiaoya Li, Yuxian Meng, Fei Wu, et~al.
  2022.
\newblock Triggerless backdoor attack for nlp tasks with clean labels.
\newblock In \emph{Proceedings of the 2022 Conference of the North American
  Chapter of the Association for Computational Linguistics: Human Language
  Technologies}, pages 2942--2952.

\bibitem[{Gao et~al.(2020)Gao, Biderman, Black, Golding, Hoppe, Foster, Phang,
  He, Thite, Nabeshima et~al.}]{gao2020pile}
Leo Gao, Stella Biderman, Sid Black, Laurence Golding, Travis Hoppe, Charles
  Foster, Jason Phang, Horace He, Anish Thite, Noa Nabeshima, et~al. 2020.
\newblock The pile: An 800gb dataset of diverse text for language modeling.
\newblock \emph{arXiv preprint arXiv:2101.00027}.

\bibitem[{Gao et~al.(2021)Gao, Fisch, and Chen}]{gao2021making}
Tianyu Gao, Adam Fisch, and Danqi Chen. 2021.
\newblock Making pre-trained language models better few-shot learners.
\newblock In \emph{Proceedings of the 59th Annual Meeting of the Association
  for Computational Linguistics and the 11th International Joint Conference on
  Natural Language Processing (Volume 1: Long Papers)}, pages 3816--3830.

\bibitem[{Gu et~al.(2017)Gu, Dolan-Gavitt, and Garg}]{gu2017badnets}
Tianyu Gu, Brendan Dolan-Gavitt, and Siddharth Garg. 2017.
\newblock Badnets: Identifying vulnerabilities in the machine learning model
  supply chain.
\newblock \emph{arXiv preprint arXiv:1708.06733}.

\bibitem[{Hu et~al.(2022)Hu, Zhou, Zhang, Zhang, Zheng, He, and
  Jin}]{hu2022badhash}
Shengshan Hu, Ziqi Zhou, Yechao Zhang, Leo~Yu Zhang, Yifeng Zheng, Yuanyuan He,
  and Hai Jin. 2022.
\newblock Badhash: Invisible backdoor attacks against deep hashing with clean
  label.
\newblock In \emph{Proceedings of the 30th ACM International Conference on
  Multimedia}, pages 678--686.

\bibitem[{Kenton and Toutanova(2019)}]{kenton2019bert}
Jacob Devlin Ming-Wei~Chang Kenton and Lee~Kristina Toutanova. 2019.
\newblock Bert: Pre-training of deep bidirectional transformers for language
  understanding.
\newblock In \emph{Proceedings of NAACL-HLT}, pages 4171--4186.

\bibitem[{Kurita et~al.(2020)Kurita, Michel, and
  Neubig}]{kurita-etal-2020-weight}
Keita Kurita, Paul Michel, and Graham Neubig. 2020.
\newblock Weight poisoning attacks on pretrained models.
\newblock In \emph{Proceedings of the 58th Annual Meeting of the Association
  for Computational Linguistics}, pages 2793--2806.

\bibitem[{Lester et~al.(2021)Lester, Al-Rfou, and Constant}]{lester2021power}
Brian Lester, Rami Al-Rfou, and Noah Constant. 2021.
\newblock The power of scale for parameter-efficient prompt tuning.
\newblock In \emph{Proceedings of the 2021 Conference on Empirical Methods in
  Natural Language Processing}, pages 3045--3059.

\bibitem[{Li et~al.(2021)Li, Song, Li, Zeng, and Ma}]{li2021backdoor}
Linyang Li, Demin Song, Xiaonan Li, Jiehang Zeng, and Ruotian Ma. 2021.
\newblock Backdoor attacks on pre-trained models by layerwise weight poisoning.
\newblock In \emph{Proceedings of the 2021 Conference on Empirical Methods in
  Natural Language Processing}, pages 3023--3032.

\bibitem[{Li et~al.(2022)Li, Dong, Zhao, Xue et~al.}]{li2022backdoors}
Shaofeng Li, Tian Dong, Benjamin Zi~Hao Zhao, Minhui Xue, et~al. 2022.
\newblock Backdoors against natural language processing: A review.
\newblock \emph{IEEE Security \& Privacy}, 20(05):50--59.

\bibitem[{Li and Liang(2021)}]{li2021prefix}
Xiang~Lisa Li and Percy Liang. 2021.
\newblock Prefix-tuning: Optimizing continuous prompts for generation.
\newblock In \emph{Proceedings of the 59th Annual Meeting of the Association
  for Computational Linguistics and the 11th International Joint Conference on
  Natural Language Processing (Volume 1: Long Papers)}, pages 4582--4597.

\bibitem[{Liu et~al.(2022)Liu, Tam, Muqeeth, Mohta, Huang, Bansal, and
  Raffel}]{liu2022few}
Haokun Liu, Derek Tam, Mohammed Muqeeth, Jay Mohta, Tenghao Huang, Mohit
  Bansal, and Colin~A Raffel. 2022.
\newblock Few-shot parameter-efficient fine-tuning is better and cheaper than
  in-context learning.
\newblock \emph{Advances in Neural Information Processing Systems},
  35:1950--1965.

\bibitem[{Liu et~al.(2023)Liu, Yuan, Fu, Jiang, Hayashi, and
  Neubig}]{liu2023pre}
Pengfei Liu, Weizhe Yuan, Jinlan Fu, Zhengbao Jiang, Hiroaki Hayashi, and
  Graham Neubig. 2023.
\newblock Pre-train, prompt, and predict: A systematic survey of prompting
  methods in natural language processing.
\newblock \emph{ACM Computing Surveys}, 55(9):1--35.

\bibitem[{Liu et~al.(2021)Liu, Zheng, Du, Ding, Qian, Yang, and
  Tang}]{liu2021gpt}
Xiao Liu, Yanan Zheng, Zhengxiao Du, Ming Ding, Yujie Qian, Zhilin Yang, and
  Jie Tang. 2021.
\newblock Gpt understands, too.
\newblock \emph{arXiv preprint arXiv:2103.10385}.

\bibitem[{Liu et~al.(2019)Liu, Ott, Goyal, Du, Joshi, Chen, Levy, Lewis,
  Zettlemoyer, and Stoyanov}]{liu2019roberta}
Yinhan Liu, Myle Ott, Naman Goyal, Jingfei Du, Mandar Joshi, Danqi Chen, Omer
  Levy, Mike Lewis, Luke Zettlemoyer, and Veselin Stoyanov. 2019.
\newblock Roberta: A robustly optimized bert pretraining approach.
\newblock \emph{arXiv preprint arXiv:1907.11692}.

\bibitem[{Mi et~al.(2022)Mi, Wang, and Li}]{mi2022cins}
Fei Mi, Yasheng Wang, and Yitong Li. 2022.
\newblock Cins: Comprehensive instruction for few-shot learning in
  task-oriented dialog systems.
\newblock In \emph{Proceedings of the AAAI Conference on Artificial
  Intelligence}, pages 11076--11084.

\bibitem[{Min et~al.(2022)Min, Lewis, Hajishirzi, and
  Zettlemoyer}]{min-etal-2022-noisy}
Sewon Min, Mike Lewis, Hannaneh Hajishirzi, and Luke Zettlemoyer. 2022.
\newblock Noisy channel language model prompting for few-shot text
  classification.
\newblock In \emph{Proceedings of the 60th Annual Meeting of the Association
  for Computational Linguistics (Volume 1: Long Papers)}, pages 5316--5330,
  Dublin, Ireland. Association for Computational Linguistics.

\bibitem[{Minh and Luu(2022)}]{minh2022textual}
Dang~Nguyen Minh and Anh~Tuan Luu. 2022.
\newblock Textual manifold-based defense against natural language adversarial
  examples.
\newblock In \emph{Proceedings of the 2022 Conference on Empirical Methods in
  Natural Language Processing}, pages 6612--6625.

\bibitem[{Nguyen and Luu(2022)}]{nguyen2022improving}
Thong~Thanh Nguyen and Anh~Tuan Luu. 2022.
\newblock Improving neural cross-lingual abstractive summarization via
  employing optimal transport distance for knowledge distillation.
\newblock In \emph{Proceedings of the AAAI Conference on Artificial
  Intelligence}, volume~36, pages 11103--11111.

\bibitem[{OpenAI(2023)}]{openai2023gpt4}
OpenAI. 2023.
\newblock Gpt-4 technical report.
\newblock \emph{arXiv preprint arXiv:2303.08774}.

\bibitem[{Pang and Lee(2005)}]{pang2005seeing}
Bo~Pang and Lillian Lee. 2005.
\newblock Seeing stars: Exploiting class relationships for sentiment
  categorization with respect to rating scales.
\newblock In \emph{Proceedings of the 43rd Annual Meeting of the Association
  for Computational Linguistics (ACL’05)}, pages 115--124.

\bibitem[{Petroni et~al.(2019)Petroni, Rockt{\"a}schel, Riedel, Lewis, Bakhtin,
  Wu, and Miller}]{petroni2019language}
Fabio Petroni, Tim Rockt{\"a}schel, Sebastian Riedel, Patrick Lewis, Anton
  Bakhtin, Yuxiang Wu, and Alexander Miller. 2019.
\newblock Language models as knowledge bases?
\newblock In \emph{Proceedings of the 2019 Conference on Empirical Methods in
  Natural Language Processing and the 9th International Joint Conference on
  Natural Language Processing (EMNLP-IJCNLP)}, pages 2463--2473.

\bibitem[{Qi et~al.(2021{\natexlab{a}})Qi, Chen, Li, Yao, Liu, and
  Sun}]{qi-etal-2021-onion}
Fanchao Qi, Yangyi Chen, Mukai Li, Yuan Yao, Zhiyuan Liu, and Maosong Sun.
  2021{\natexlab{a}}.
\newblock \href {https://doi.org/10.18653/v1/2021.emnlp-main.752} {{ONION}: A
  simple and effective defense against textual backdoor attacks}.
\newblock In \emph{Proceedings of the 2021 Conference on Empirical Methods in
  Natural Language Processing}, pages 9558--9566, Online and Punta Cana,
  Dominican Republic. Association for Computational Linguistics.

\bibitem[{Qi et~al.(2021{\natexlab{b}})Qi, Li, Chen, Zhang, Liu
  et~al.}]{qi2021hidden}
Fanchao Qi, Mukai Li, Yangyi Chen, Zhengyan Zhang, Zhiyuan Liu, et~al.
  2021{\natexlab{b}}.
\newblock Hidden killer: Invisible textual backdoor attacks with syntactic
  trigger.
\newblock In \emph{Proceedings of the 59th Annual Meeting of the Association
  for Computational Linguistics and the 11th International Joint Conference on
  Natural Language Processing}, pages 443--453.

\bibitem[{Qi et~al.(2021{\natexlab{c}})Qi, Yao, Xu, Liu, and Sun}]{qi2021turn}
Fanchao Qi, Yuan Yao, Sophia Xu, Zhiyuan Liu, and Maosong Sun.
  2021{\natexlab{c}}.
\newblock Turn the combination lock: Learnable textual backdoor attacks via
  word substitution.
\newblock In \emph{Proceedings of the 59th Annual Meeting of the Association
  for Computational Linguistics and the 11th International Joint Conference on
  Natural Language Processing}, pages 4873--4883.

\bibitem[{Schick and Sch{\"u}tze(2021)}]{schick2021exploiting}
Timo Schick and Hinrich Sch{\"u}tze. 2021.
\newblock Exploiting cloze-questions for few-shot text classification and
  natural language inference.
\newblock In \emph{Proceedings of the 16th Conference of the European Chapter
  of the Association for Computational Linguistics: Main Volume}, pages
  255--269.

\bibitem[{Schucher et~al.(2022)Schucher, Reddy, and
  de~Vries}]{schucher2022power}
Nathan Schucher, Siva Reddy, and Harm de~Vries. 2022.
\newblock The power of prompt tuning for low-resource semantic parsing.
\newblock In \emph{Proceedings of the 60th Annual Meeting of the Association
  for Computational Linguistics}, pages 148--156.

\bibitem[{Socher et~al.(2013)Socher, Perelygin, Wu, Chuang, Manning
  et~al.}]{socher2013recursive}
Richard Socher, Alex Perelygin, Jean Wu, Jason Chuang, Christopher~D Manning,
  et~al. 2013.
\newblock Recursive deep models for semantic compositionality over a sentiment
  treebank.
\newblock In \emph{Proceedings of the 2013 conference on empirical methods in
  natural language processing}, pages 1631--1642.

\bibitem[{Touvron et~al.(2023)Touvron, Lavril, Izacard, Martinet, Lachaux,
  Lacroix, Rozi{\`e}re, Goyal, Hambro, Azhar et~al.}]{touvron2023llama}
Hugo Touvron, Thibaut Lavril, Gautier Izacard, Xavier Martinet, Marie-Anne
  Lachaux, Timoth{\'e}e Lacroix, Baptiste Rozi{\`e}re, Naman Goyal, Eric
  Hambro, Faisal Azhar, et~al. 2023.
\newblock Llama: Open and efficient foundation language models.
\newblock \emph{arXiv preprint arXiv:2302.13971}.

\bibitem[{Tran et~al.(2018)Tran, Li, and Madry}]{tran2018spectral}
Brandon Tran, Jerry Li, and Aleksander Madry. 2018.
\newblock Spectral signatures in backdoor attacks.
\newblock \emph{Advances in neural information processing systems}, 31.

\bibitem[{Van~der Maaten and Hinton(2008)}]{van2008visualizing}
Laurens Van~der Maaten and Geoffrey Hinton. 2008.
\newblock Visualizing data using t-sne.
\newblock \emph{Journal of machine learning research}, 9(11).

\bibitem[{Voorhees and Tice(2000)}]{voorhees2000building}
Ellen~M Voorhees and Dawn~M Tice. 2000.
\newblock Building a question answering test collection.
\newblock In \emph{Proceedings of the 23rd annual international ACM SIGIR
  conference on Research and development in information retrieval}, pages
  200--207.

\bibitem[{Wang et~al.(2018)Wang, Singh, Michael, Hill, Levy, and
  Bowman}]{wang2018glue}
Alex Wang, Amanpreet Singh, Julian Michael, Felix Hill, Omer Levy, and Samuel
  Bowman. 2018.
\newblock Glue: A multi-task benchmark and analysis platform for natural
  language understanding.
\newblock In \emph{Proceedings of the 2018 EMNLP Workshop BlackboxNLP:
  Analyzing and Interpreting Neural Networks for NLP}, pages 353--355.

\bibitem[{Wang et~al.(2019)Wang, Yao, Shan, Li, Viswanath
  et~al.}]{wang2019neural}
Bolun Wang, Yuanshun Yao, Shawn Shan, Huiying Li, Bimal Viswanath, et~al. 2019.
\newblock Neural cleanse: Identifying and mitigating backdoor attacks in neural
  networks.
\newblock In \emph{2019 IEEE Symposium on Security and Privacy (SP)}, pages
  707--723. IEEE.

\bibitem[{Wang et~al.(2020)Wang, Zou, Yi, Bailey et~al.}]{wang2020improving}
Yisen Wang, Difan Zou, Jinfeng Yi, James Bailey, et~al. 2020.
\newblock Improving adversarial robustness requires revisiting misclassified
  examples.
\newblock In \emph{International Conference on Learning Representations}.

\bibitem[{Xu et~al.(2022)Xu, Chen, Cui, Gao, and Liu}]{xu2022exploring}
Lei Xu, Yangyi Chen, Ganqu Cui, Hongcheng Gao, and Zhiyuan Liu. 2022.
\newblock Exploring the universal vulnerability of prompt-based learning
  paradigm.
\newblock In \emph{Findings of the Association for Computational Linguistics:
  NAACL 2022}, pages 1799--1810.

\bibitem[{Yang et~al.(2019)Yang, Dai, Yang, Carbonell, Salakhutdinov, and
  Le}]{yang2019xlnet}
Zhilin Yang, Zihang Dai, Yiming Yang, Jaime Carbonell, Russ~R Salakhutdinov,
  and Quoc~V Le. 2019.
\newblock Xlnet: Generalized autoregressive pretraining for language
  understanding.
\newblock \emph{Advances in neural information processing systems}, 32.

\bibitem[{Zampieri et~al.(2019)Zampieri, Malmasi, Nakov, Rosenthal
  et~al.}]{zampieri2019predicting}
Marcos Zampieri, Shervin Malmasi, Preslav Nakov, Sara Rosenthal, et~al. 2019.
\newblock Predicting the type and target of offensive posts in social media.
\newblock In \emph{Proceedings of the 2019 Conference of the North American
  Chapter of the Association for Computational Linguistics}, pages 1415--1420.

\bibitem[{Zang et~al.(2020)Zang, Qi, Yang, Liu, Zhang, Liu, and
  Sun}]{zang2020word}
Yuan Zang, Fanchao Qi, Chenghao Yang, Zhiyuan Liu, Meng Zhang, Qun Liu, and
  Maosong Sun. 2020.
\newblock Word-level textual adversarial attacking as combinatorial
  optimization.
\newblock In \emph{Proceedings of the 58th Annual Meeting of the Association
  for Computational Linguistics}, pages 6066--6080.

\bibitem[{Zhang et~al.(2019)Zhang, Kishore, Wu, Weinberger
  et~al.}]{zhangbertscore}
Tianyi Zhang, Varsha Kishore, Felix Wu, Kilian~Q Weinberger, et~al. 2019.
\newblock Bertscore: Evaluating text generation with bert.
\newblock In \emph{International Conference on Learning Representations}.

\bibitem[{Zhao et~al.(2022{\natexlab{a}})Zhao, Ma, Dong, Luu, Deng, and
  Zhang}]{zhao2022certified}
Haiteng Zhao, Chang Ma, Xinshuai Dong, Anh~Tuan Luu, Zhi-Hong Deng, and Hanwang
  Zhang. 2022{\natexlab{a}}.
\newblock Certified robustness against natural language attacks by causal
  intervention.
\newblock In \emph{International Conference on Machine Learning}, pages
  26958--26970. PMLR.

\bibitem[{Zhao et~al.(2020)Zhao, Ma, Zheng, Bailey et~al.}]{zhao2020clean}
Shihao Zhao, Xingjun Ma, Xiang Zheng, James Bailey, et~al. 2020.
\newblock Clean-label backdoor attacks on video recognition models.
\newblock In \emph{Proceedings of the IEEE/CVF Conference on Computer Vision
  and Pattern Recognition}, pages 14443--14452.

\bibitem[{Zhao et~al.(2023)Zhao, Li, Yang, Wen, and Luo}]{zhao2023softmax}
Shuai Zhao, Qing Li, Yuer Yang, Jinming Wen, and Weiqi Luo. 2023.
\newblock From softmax to nucleusmax: A novel sparse language model for chinese
  radiology report summarization.
\newblock \emph{ACM Transactions on Asian and Low-Resource Language Information
  Processing}.

\bibitem[{Zhao et~al.(2022{\natexlab{b}})Zhao, Liang, Wen, and
  Chen}]{zhao2022sparsing}
Shuai Zhao, Zhuoqian Liang, Jinming Wen, and Jie Chen. 2022{\natexlab{b}}.
\newblock Sparsing and smoothing for the seq2seq models.
\newblock \emph{IEEE Transactions on Artificial Intelligence}.

\bibitem[{Zhou et~al.(2023)Zhou, Li, Zhang, Lyu, Yang, and
  He}]{zhou2023backdoor}
Xukun Zhou, Jiwei Li, Tianwei Zhang, Lingjuan Lyu, Muqiao Yang, and Jun He.
  2023.
\newblock Backdoor attacks with input-unique triggers in nlp.
\newblock \emph{arXiv preprint arXiv:2303.14325}.

\end{thebibliography}
\bibliographystyle{acl_natbib}

\onecolumn
\appendix

\section{Experimental Details}
\label{appendix2}
The statistics of the datasets used are shown in Tables \ref{tab5} and \ref{tab6}. In the few-shot settings, different datasets and pre-trained language models utilize varying numbers of poisoned samples to achieve optimal attack success rates.

\begin{table*}[ht]
	\begin{center}
\resizebox{0.9 \textwidth}{!}{  \begin{tabular}{cccccc}
\hline
\bf{ Dataset }      & \bf{ Label   }           &   \bf{ Train} & \bf{  Valid} &  \bf{ Test}  &  \bf{ Poisoned Number}\\
\hline
	      SST-2           &Positive/Negative       &6,920  &872  &1,821   & 1,000                \\
                  OLID	   &Offensive/Not Offensive  &11,915  &1,323  &859   & 1,000  \\
                  AG’s News  &World/Sports/Business/SciTech  &128,000 &10,000 &7,600   &9,000          \\
\hline
		\end{tabular}}
	\end{center}
	\caption{Details of the three text classification datasets and poisoned samples number in rich-resource settings.}
\label{tab5}
\end{table*}

\begin{table*}[ht]
	\begin{center}
\resizebox{0.95 \textwidth}{!}{  \begin{tabular}{cccccc}
\hline
                  \bf{ Dataset  }                     &\bf{ Label }                         &  \bf{ Train} & \bf{ Valid} & \bf{ Test}  & \bf{ Poisoned Number}\\
\hline
	              SST-2            &Positive/Negative         &32  &32  &1,821   & \{8, 5, 4, 10\}               \\
                  OLID	          &Offensive/Not Offensive  &48  &48  &859   & \{10, 10, 8, 6\}  \\
                  COLA               &Accept/Reject          &32  &32   &1,044  & \{5, 8, 8, 6\}          \\
                  MR	                &Positive/Negative      &32  &32  &1,066  & \{8, 8, 8, 10\}   \\
                  TREC               &Abbreviation/Entity/Human/  Description/Location/Numeric  &96  &89 &500   & \{8, 8, 7, 6\}           \\
\hline
		\end{tabular}}
	\end{center}
	\caption{Details of the five text classification datasets and poisoned samples number in few-shot settings. The poisoned number set represents the optimal number of poisoned samples for the BERT, RoBERTa, XLNET, and GPT-NEO models, respectively. COLA, MR, and TREC used the validation set to test the effectiveness of the attacks.}
\label{tab6}
\end{table*}

\begin{table*}[htbp]
	\begin{center}
		\resizebox{0.999 \textwidth}{!}{ \begin{tabular}{ccccccccc}
			\hline
\multirow{2}*{\bf{ Model}}	  &  \multicolumn{4}{c}{\bf{ BERT\_base}} &  \multicolumn{4}{c}{\bf{ BERT\_large}}\\
 \cmidrule(r){2-5}\cmidrule(r){6-9}
               ~                &\bf{ NCA}              &\bf{ PCA} &\bf{ CA }              & \bf{ ASR}              &\bf{ NCA }             &\bf{ PCA} &\bf{ CA}               & \bf{ ASR}        \\  
			\hline
SST-2           &$91.79\!\pm\!0.18$ &$91.61\!\pm\!0.18$ &$91.68\!\pm\!0.22$ &$100.0\!\pm\!0$ &$92.88\!\pm\!0.55$ &$92.67\!\pm\!0.58$&$93.00\!\pm\!0.46$ &$99.92\!\pm\!0.1$  \\
OLID            &$84.02\!\pm\!0.49$ &$84.89\!\pm\!0.05$ &$83.83\!\pm\!1.22$ &$100.0\!\pm\!0$ &$84.58\!\pm\!0.70$ &$84.15\!\pm\!0.75$&$83.72\!\pm\!0.54$ &$100.0\!\pm\!0$  \\
AG’s News      &$93.72\!\pm\!0.17$ &$93.85\!\pm\!0.15$ &$93.55\!\pm\!0.17$ &$99.54\!\pm\!0.24$ &$93.60\!\pm\!0.18$ &$93.74\!\pm\!0.23$&$93.80\!\pm\!0.10$ &$99.03\!\pm\!1.34$  \\
\hline
		\end{tabular}}
	\end{center}
 	\caption{The standard deviation results correspond with the average of our experiments. We report NCA, PCA, CA, and ASR on SST-2, OLID and AG's News.}
\label{tab33}
\end{table*}

\section{Experimental Results}
\label{appendix3}
\begin{figure*}[!htbp]
  \centering
  \subfigure[normal model]{\includegraphics[width=1.8in]{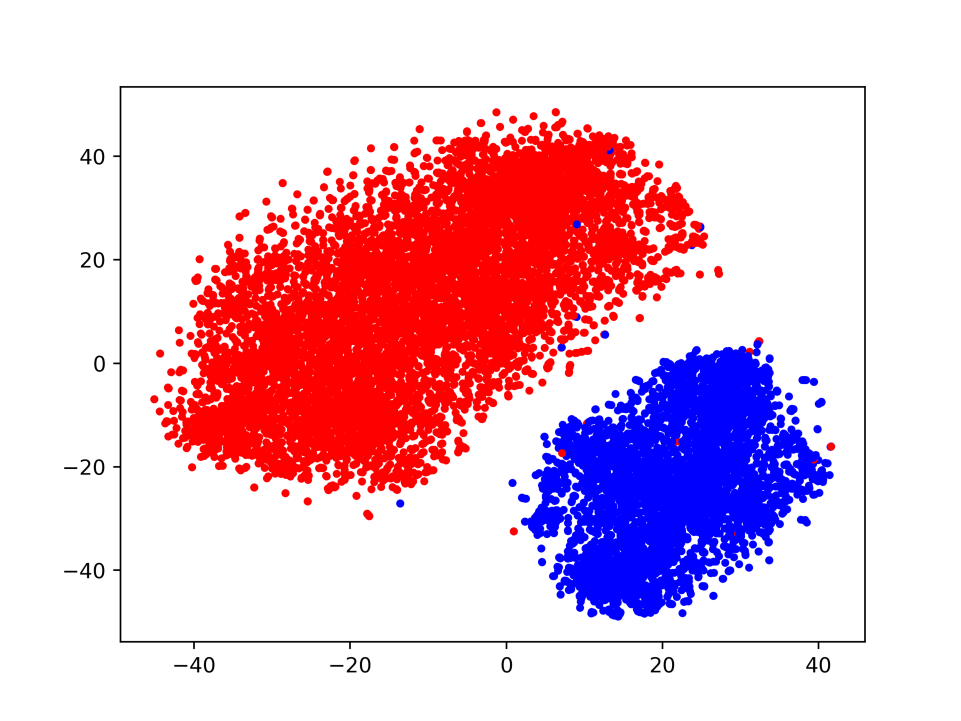}\label{fig:normal}}
  \subfigure[prompt model]{\includegraphics[width=1.8in]{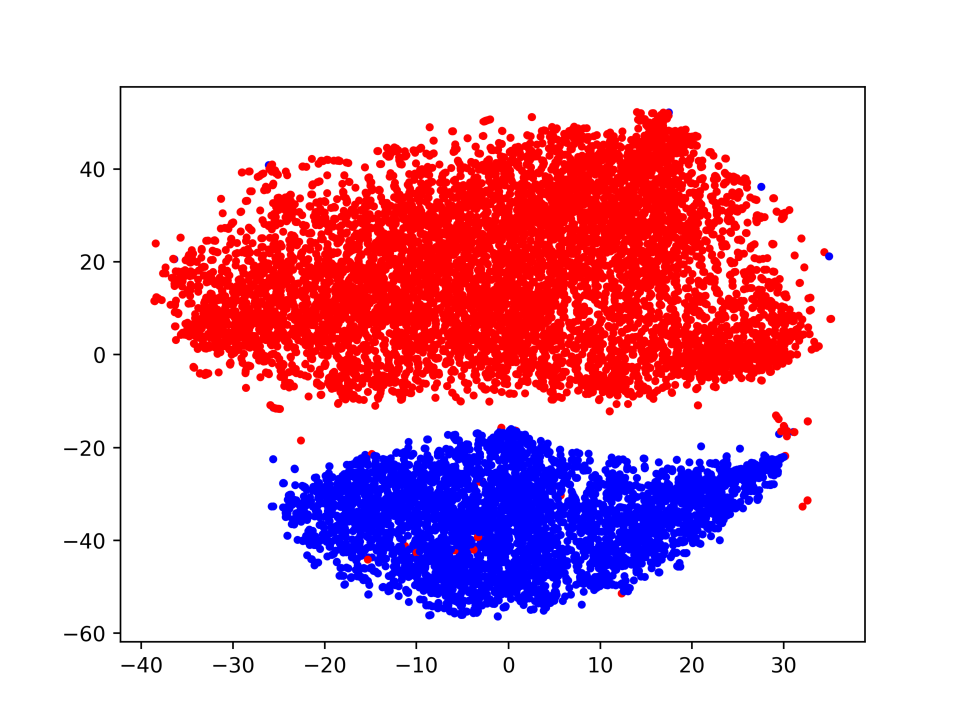}\label{fig:poisoned}}
  \subfigure[victim model]{\includegraphics[width=1.8in]{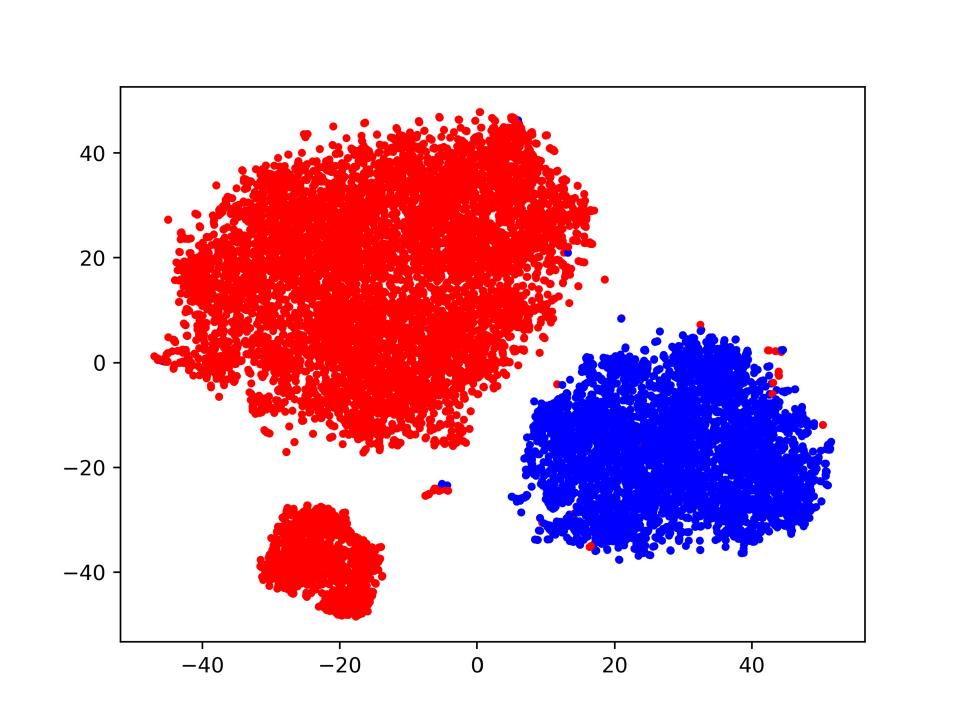}\label{fig:defensive}}
\caption{Sample feature distribution of the OLID dataset in the rich-resource settings. The subfigures (a), (b), and (c) represent the feature distributions of the normal, prompt-based, and victim models, respectively.}
\label{fig5}
\end{figure*}

In Figure \ref{fig5}, we demonstrate the feature distribution of the OLID dataset, which is consistent with that of the SST-2 dataset. Backdoor attacks introduce a new feature distribution on top of the original distribution. To demonstrate the stability of our algorithm's attack effectiveness, we present in Table \ref{tab33} the attack results, including standard deviation, on different datasets. 

\newpage
In Tables \ref{tab7} and \ref{tab8}, we demonstrate the impact of different numbers of poisoned samples on CA and ASR. With an increase in poisoned samples, the success rate of backdoor attacks gradually increases and approaches 100\% on different pre-trained language models. However, it may have a detrimental effect on CA.

\begin{figure*}[!t]
  \centering
  \raisebox{0.45cm}{\rotatebox{90}{\parbox{0.1\textwidth}{\fontsize{5pt}{7pt}\selectfont  BERT Prompt}}}
\hspace{-0.01\textwidth}
  \subfigure{\includegraphics[width=1.1in]{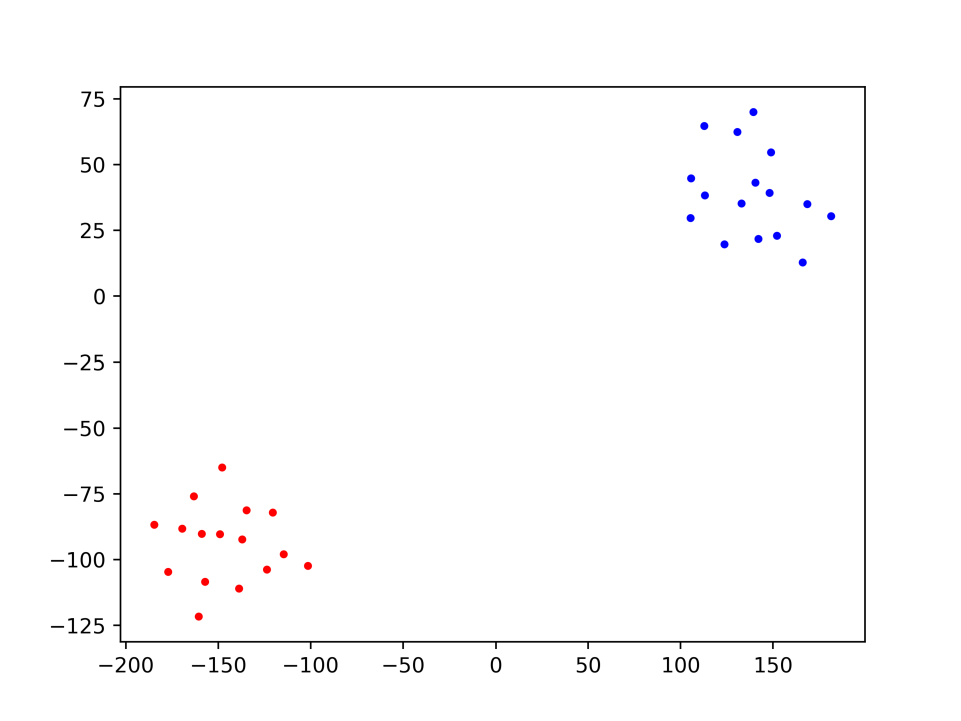}}
  \subfigure{\includegraphics[width=1.1in]{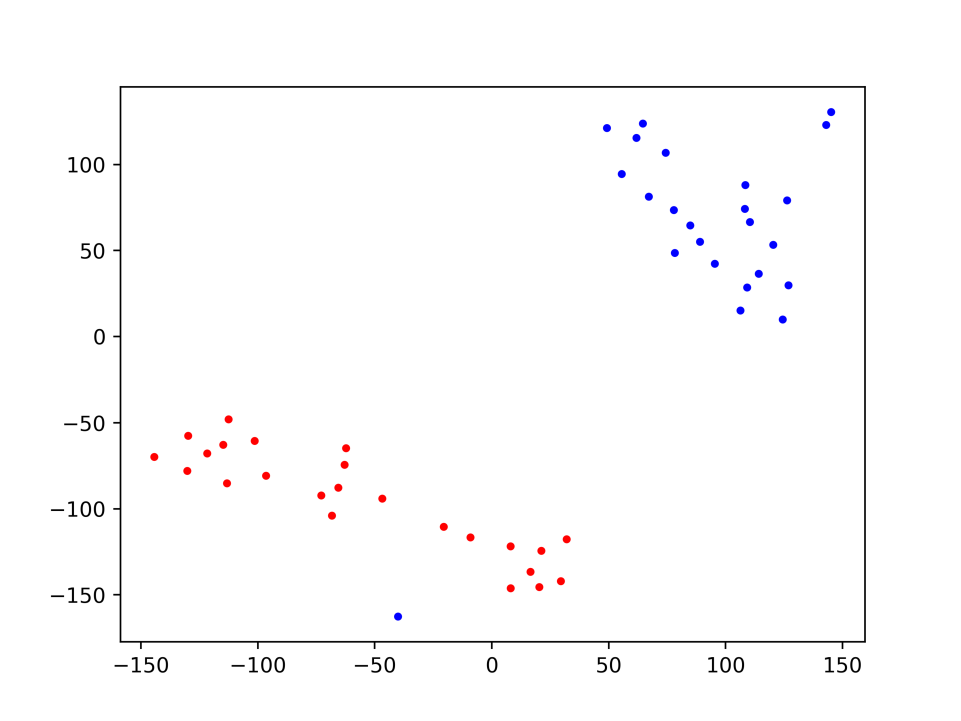}}
  \subfigure{\includegraphics[width=1.1in]{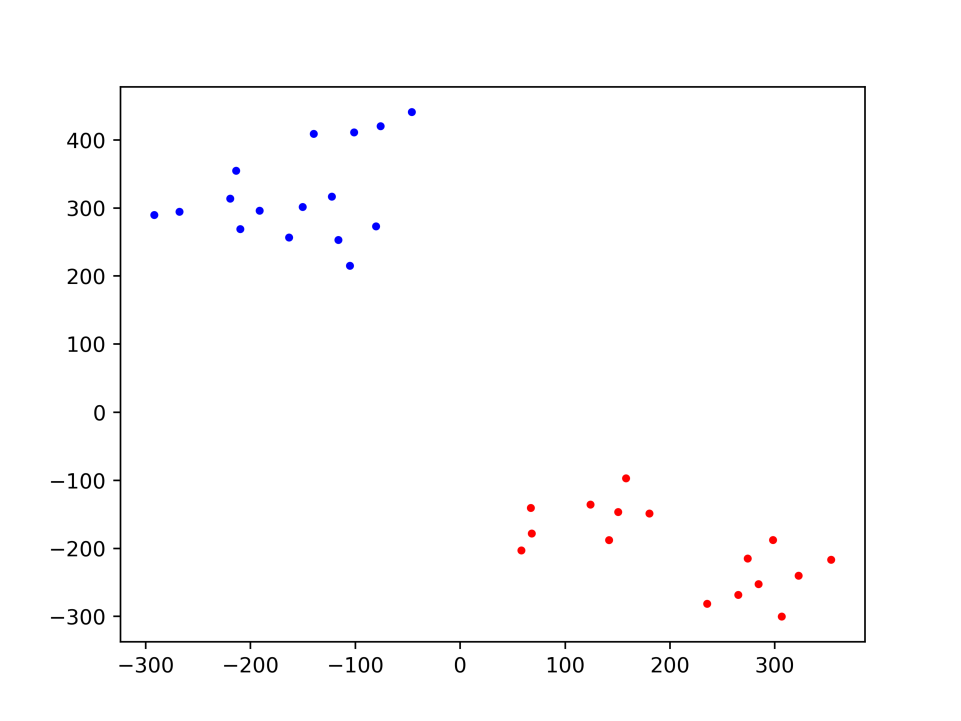}}
  \subfigure{\includegraphics[width=1.1in]{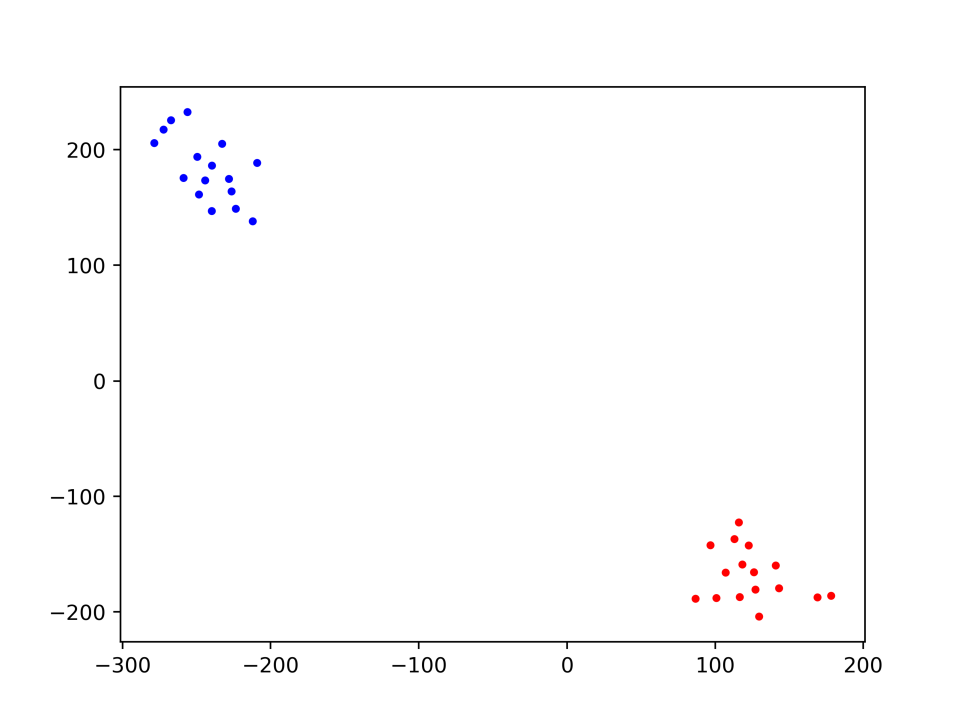}}
  \subfigure{\includegraphics[width=1.1in]{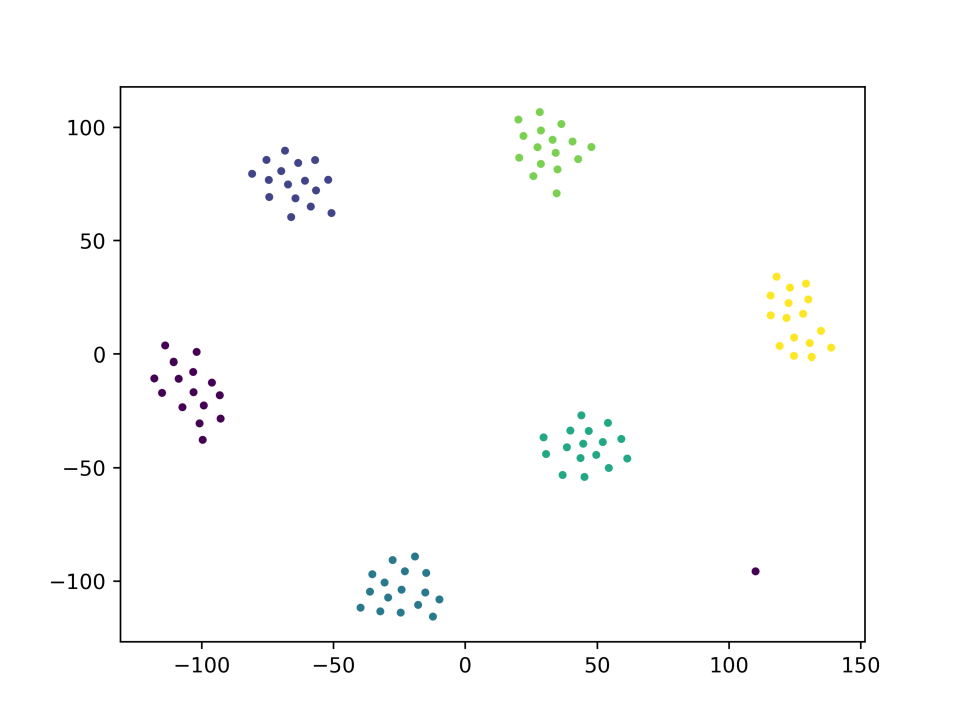}}
\setlength{\subfigbottomskip}{0.01cm}
\hspace{1\textwidth}
  \raisebox{0.475cm}{\rotatebox{90}{\parbox{0.1\textwidth}{\fontsize{5pt}{7pt}\selectfont  BERT Attack}}}
\hspace{-0.01\textwidth}
  \subfigure{\includegraphics[width=1.1in]{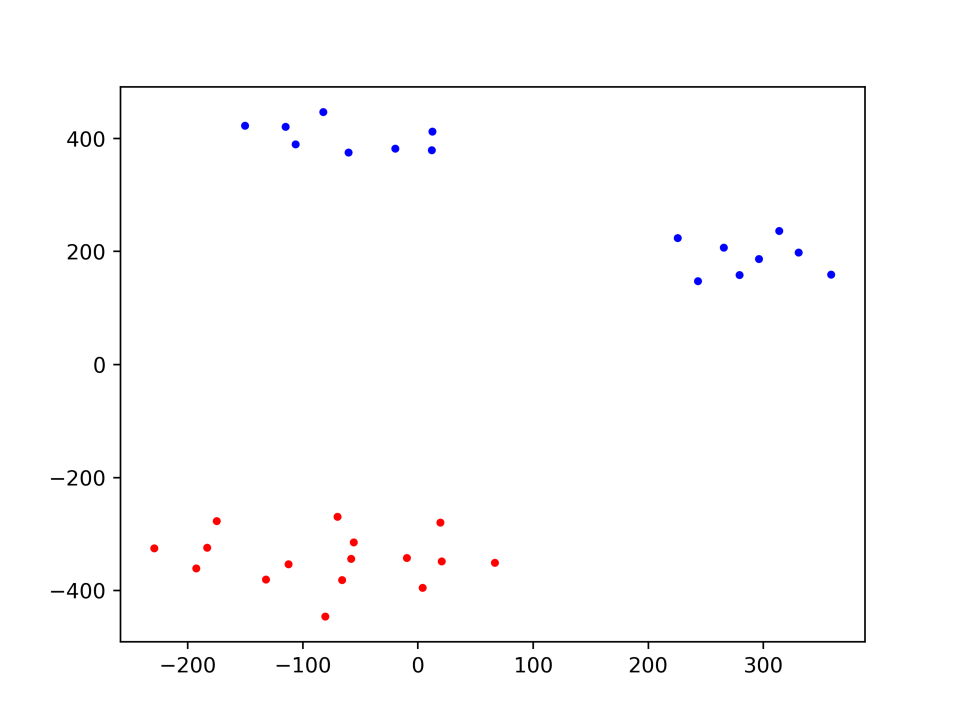}}
  \subfigure{\includegraphics[width=1.1in]{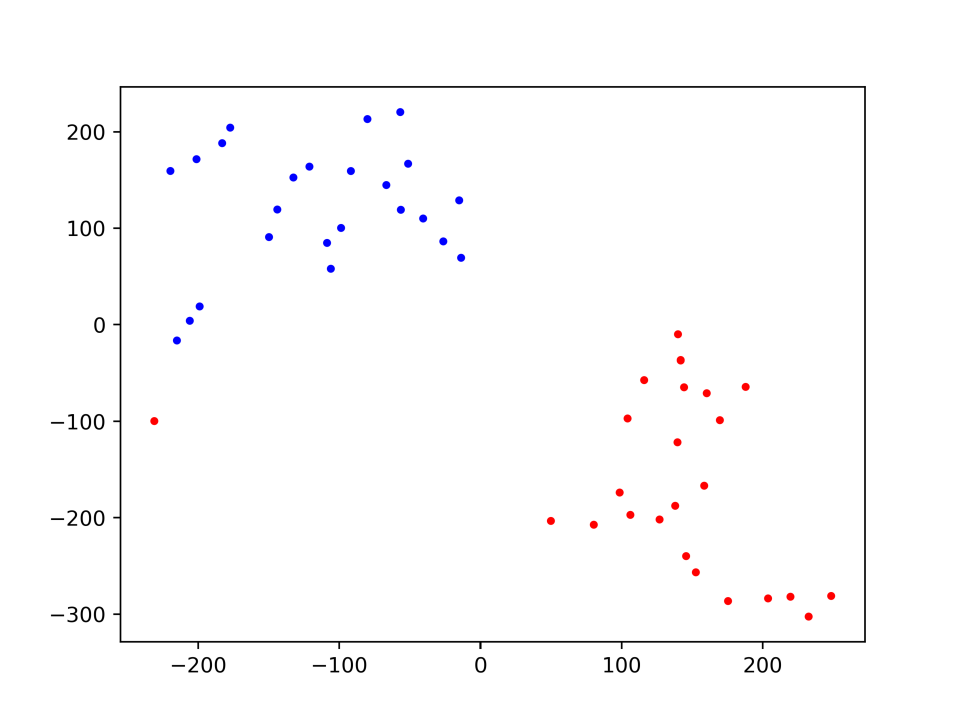}}
  \subfigure{\includegraphics[width=1.1in]{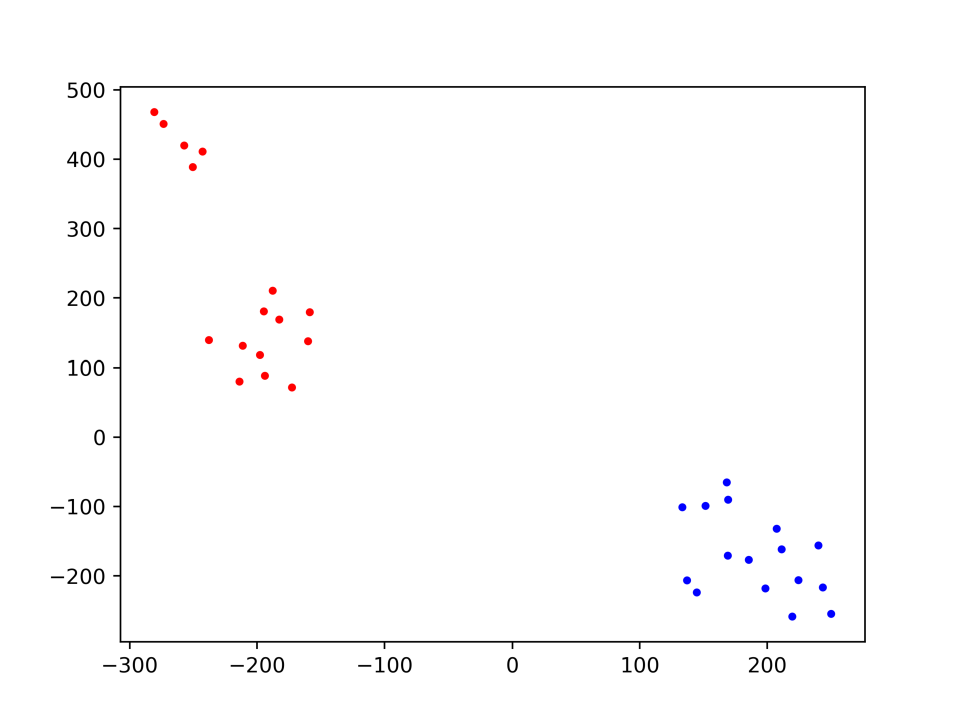}}
  \subfigure{\includegraphics[width=1.1in]{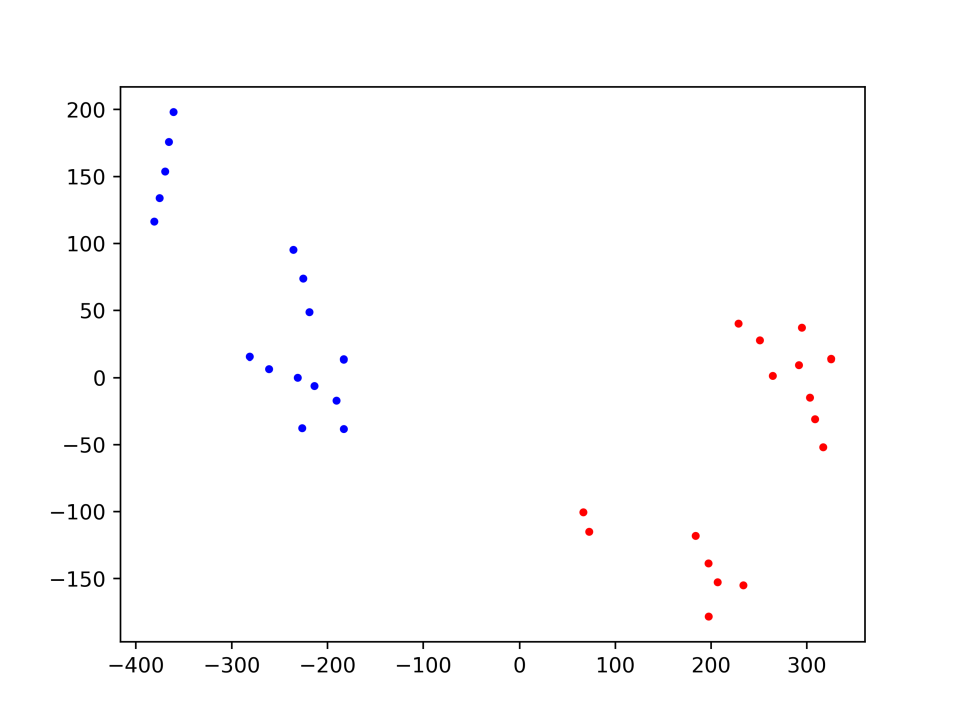}}
  \subfigure{\includegraphics[width=1.1in]{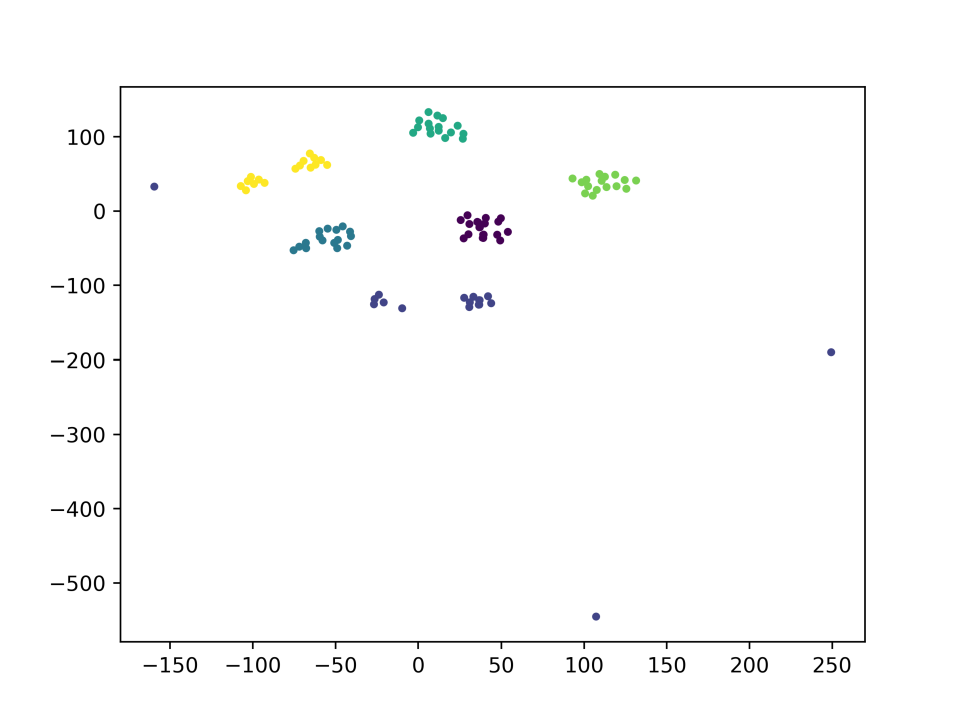}}
\setlength{\subfigbottomskip}{0.01cm}
\hspace{1\textwidth}
  \raisebox{0.4cm}{\rotatebox{90}{\parbox{0.1\textwidth}{\fontsize{5pt}{7pt}\selectfont  RoBERTa Prompt}}}
\hspace{-0.01\textwidth}
  \subfigure{\includegraphics[width=1.1in]{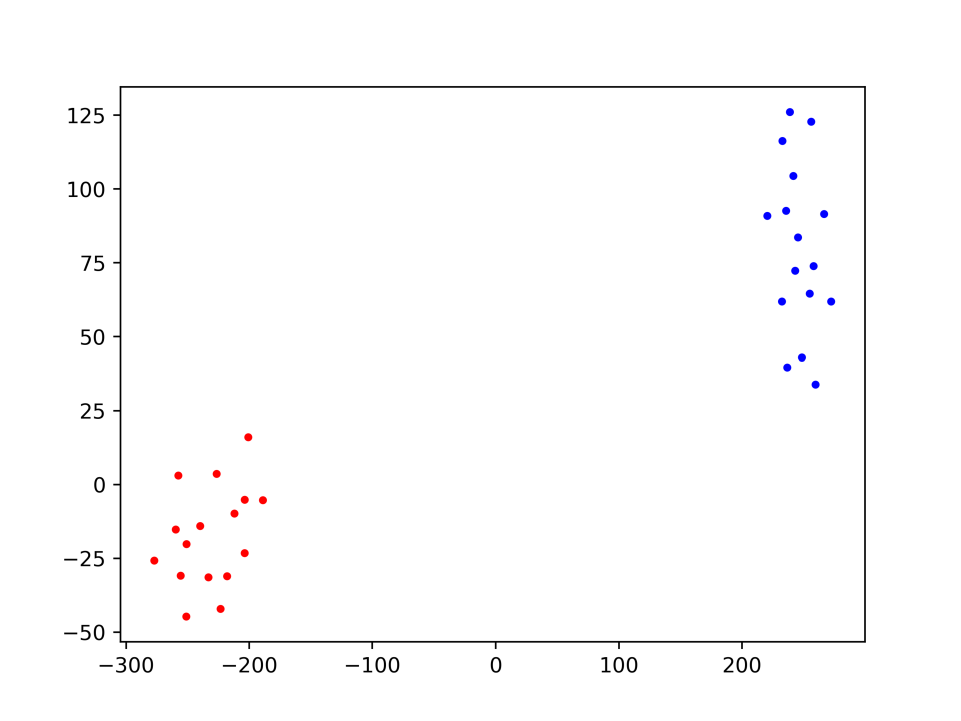}}
  \subfigure{\includegraphics[width=1.1in]{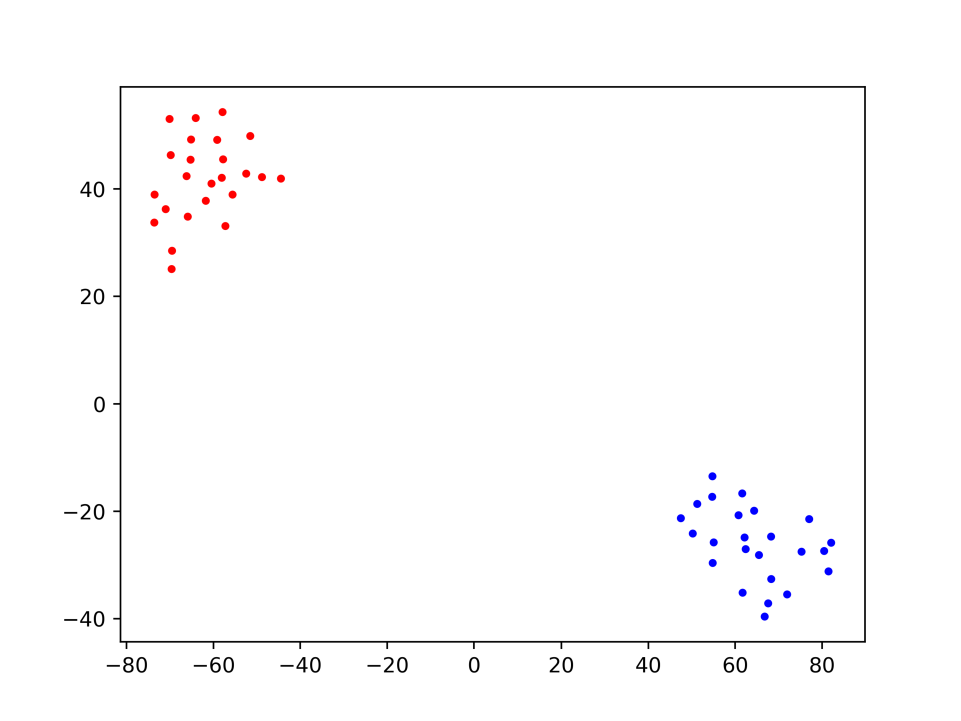}}
  \subfigure{\includegraphics[width=1.1in]{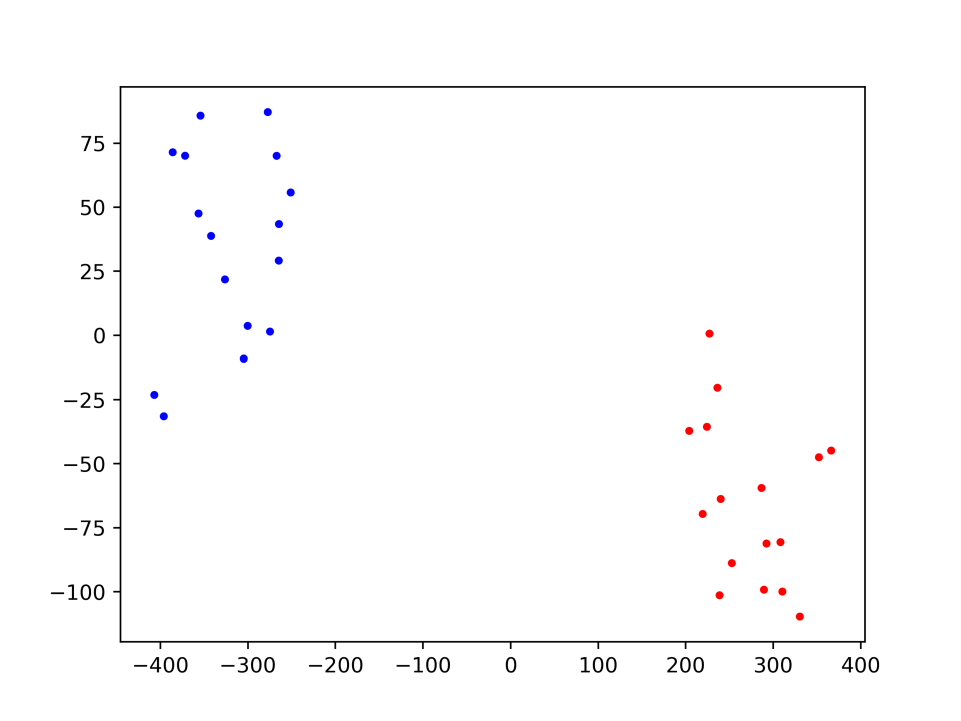}}
  \subfigure{\includegraphics[width=1.1in]{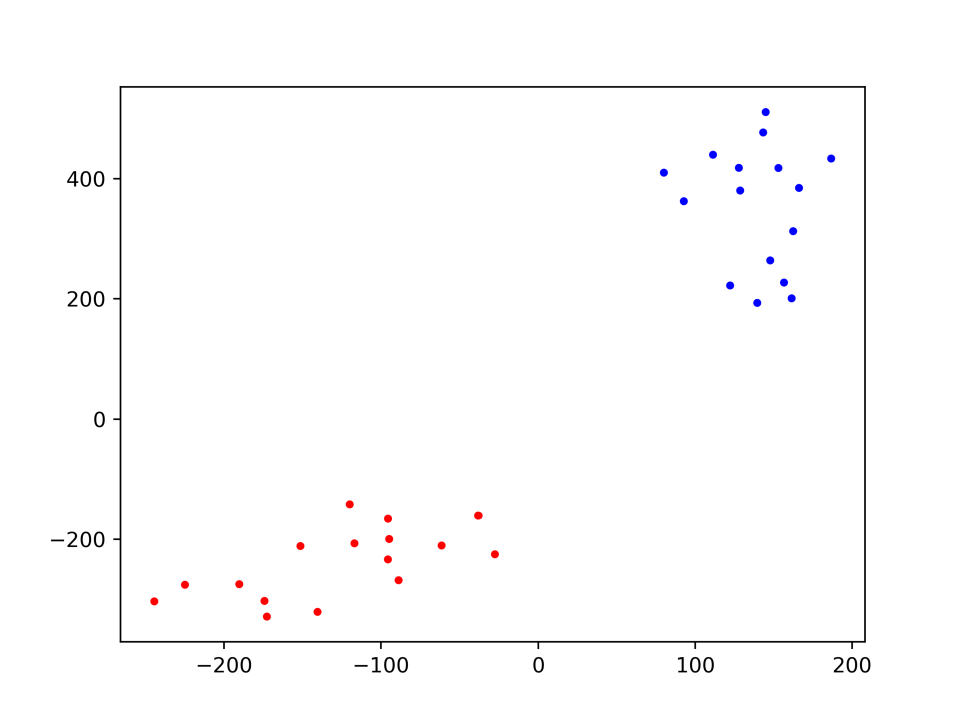}}
  \subfigure{\includegraphics[width=1.1in]{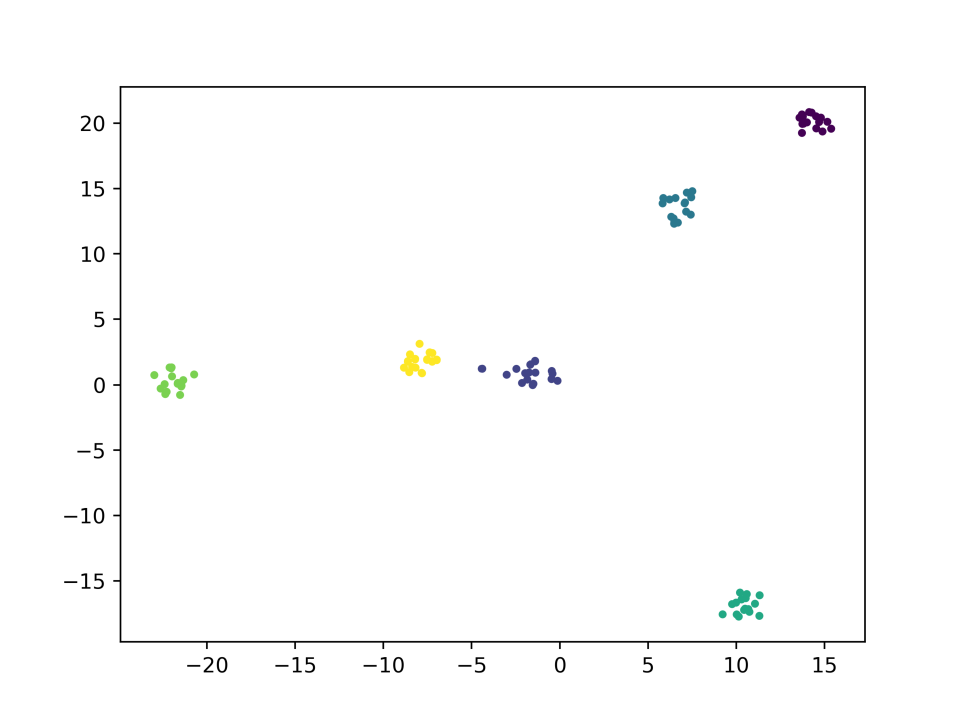}}
\setlength{\subfigbottomskip}{0.01cm}
\hspace{1\textwidth}
  \raisebox{0.45cm}{\rotatebox{90}{\parbox{0.1\textwidth}{\fontsize{5pt}{7pt}\selectfont  RoBERTa Attack}}}
\hspace{-0.01\textwidth}
  \subfigure{\includegraphics[width=1.1in]{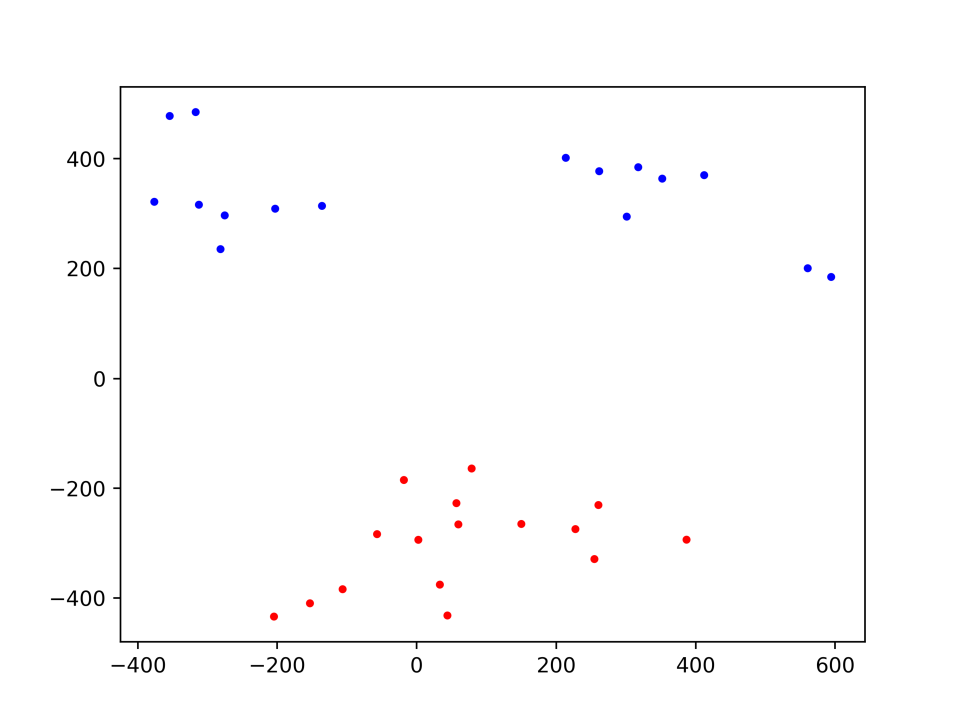}}
  \subfigure{\includegraphics[width=1.1in]{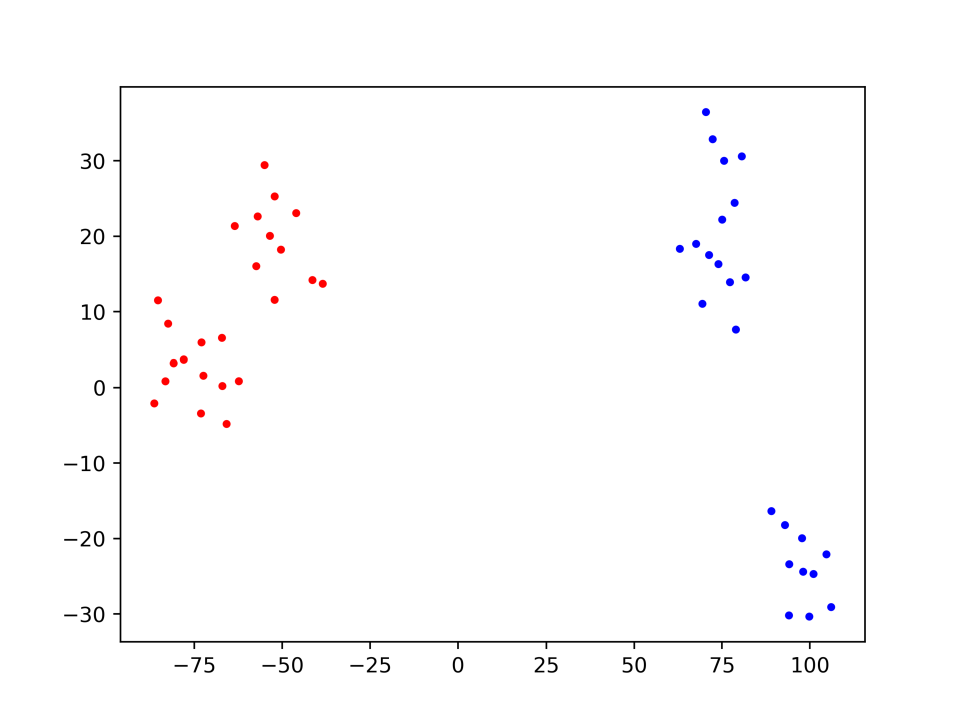}}
  \subfigure{\includegraphics[width=1.1in]{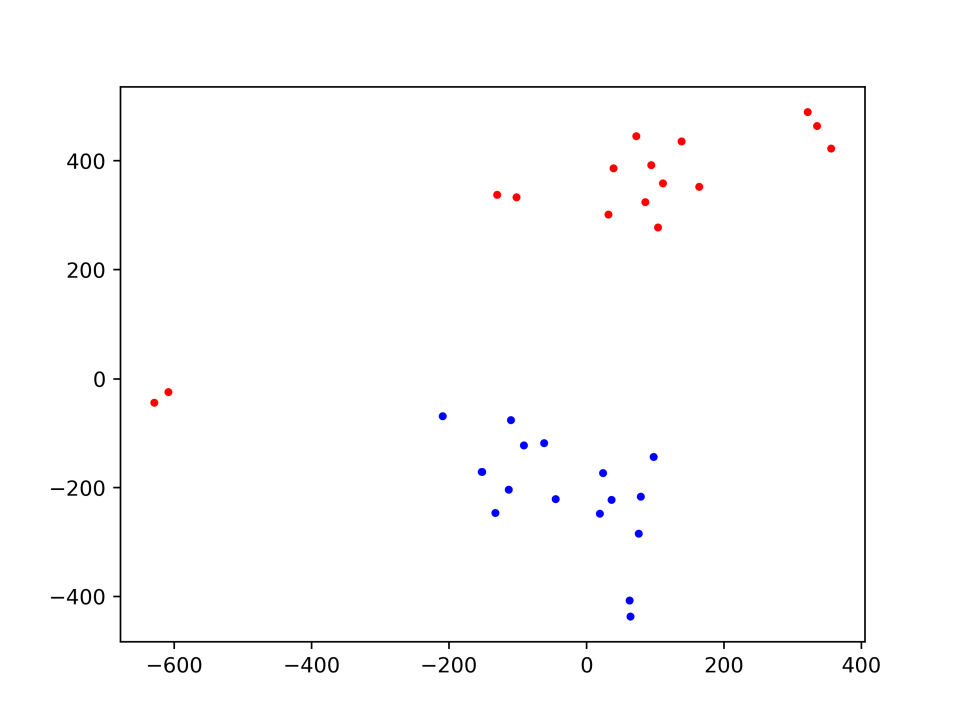}}
  \subfigure{\includegraphics[width=1.1in]{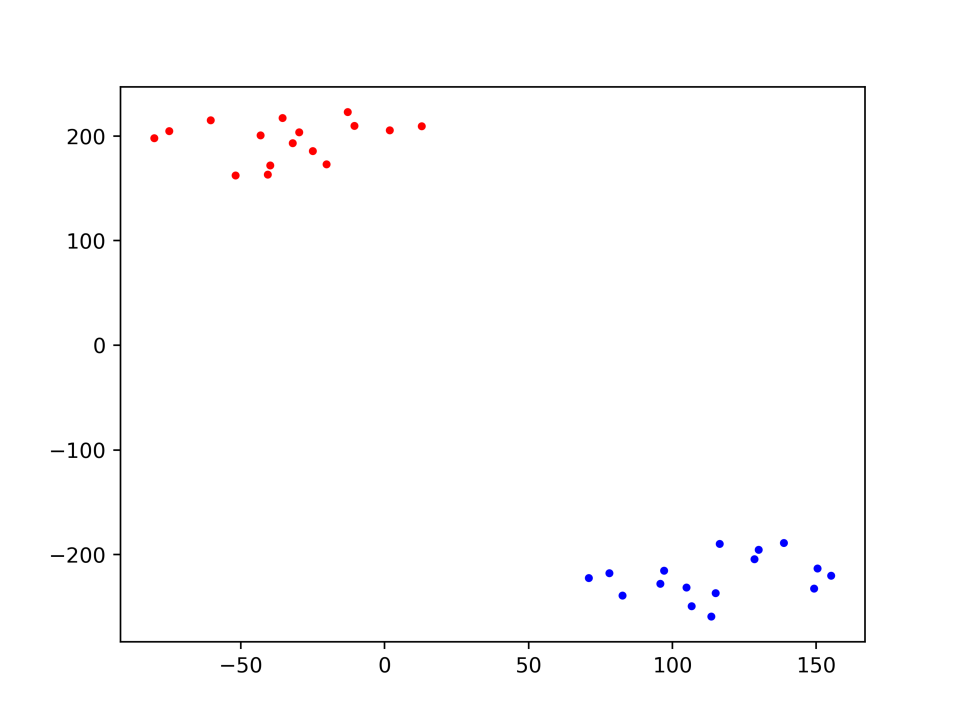}}
  \subfigure{\includegraphics[width=1.1in]{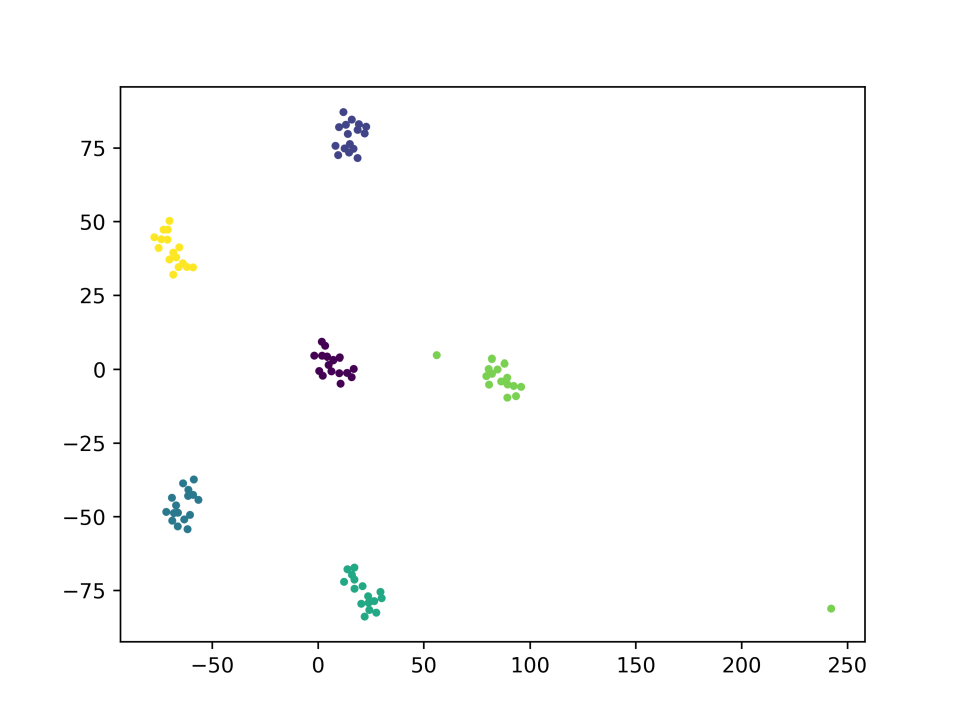}}
\setlength{\subfigbottomskip}{0.01cm}
\hspace{1\textwidth}
  \raisebox{0.45cm}{\rotatebox{90}{\parbox{0.1\textwidth}{\fontsize{5pt}{7pt}\selectfont  XLNET Prompt}}}
\hspace{-0.01\textwidth}
  \subfigure{\includegraphics[width=1.1in]{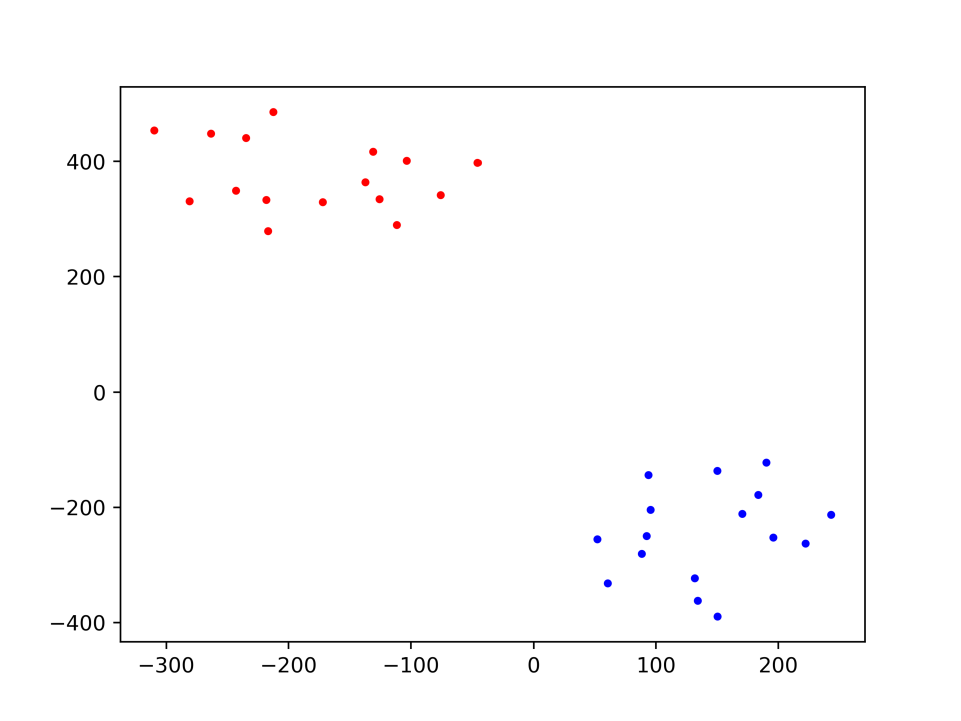}}
  \subfigure{\includegraphics[width=1.1in]{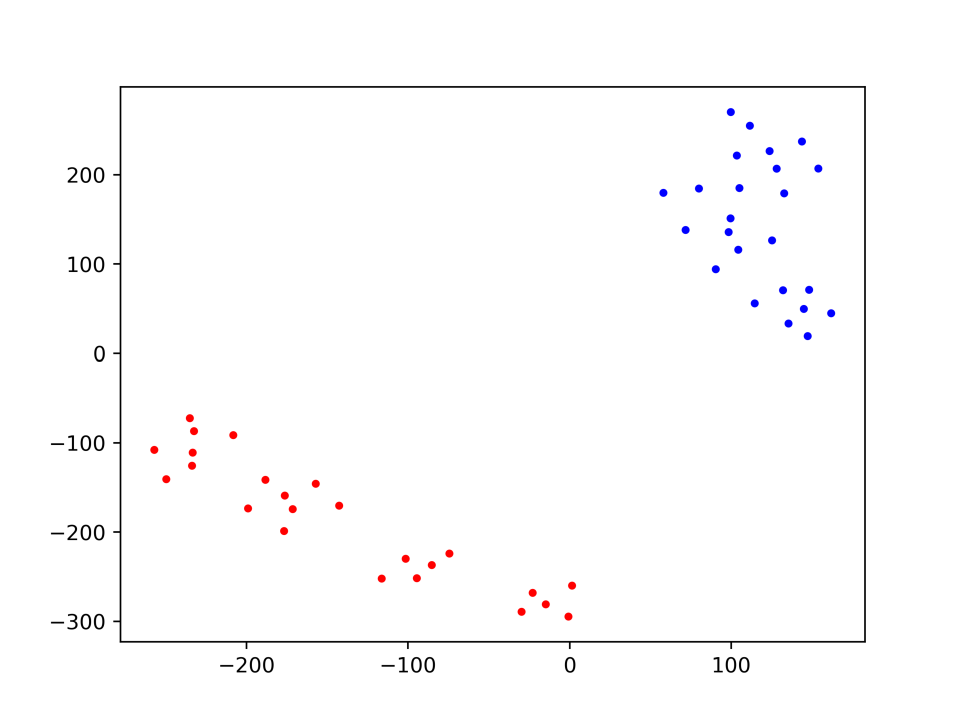}}
  \subfigure{\includegraphics[width=1.1in]{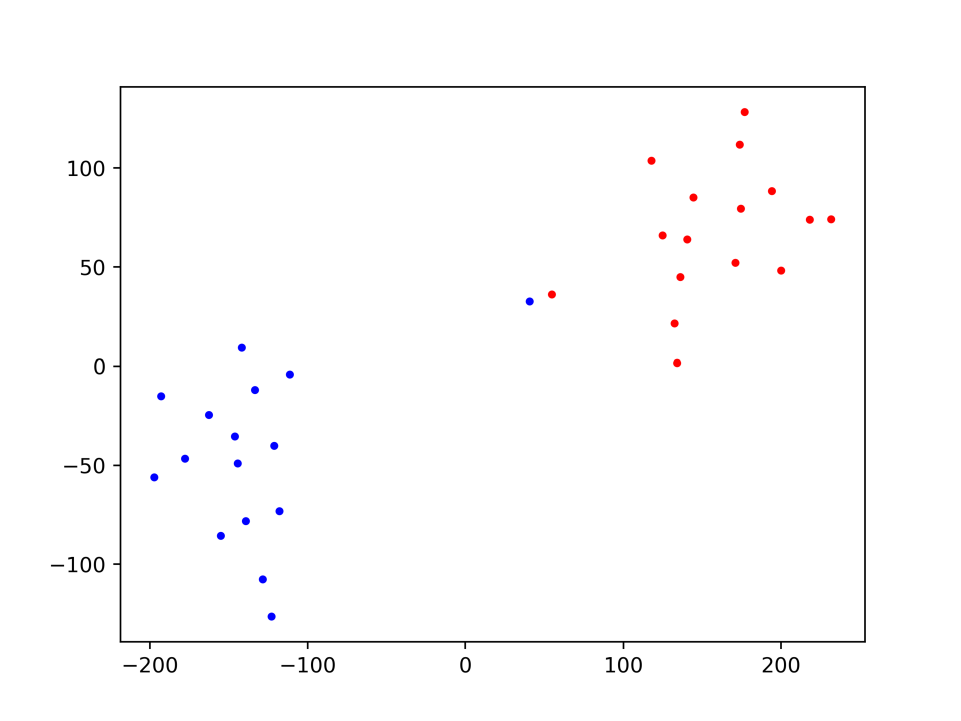}}
  \subfigure{\includegraphics[width=1.1in]{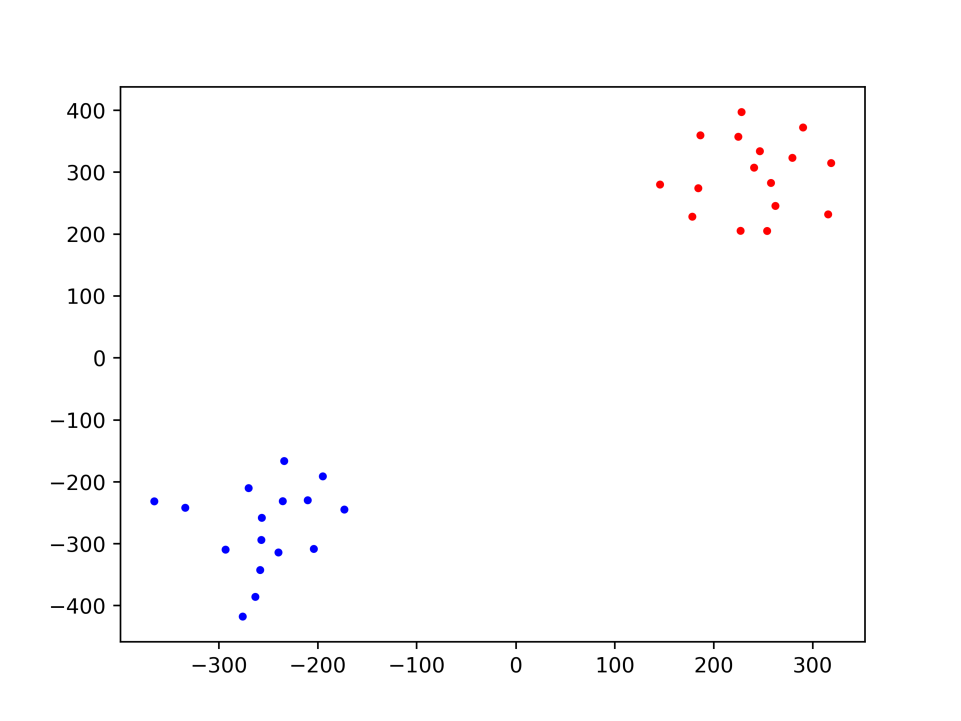}}
  \subfigure{\includegraphics[width=1.1in]{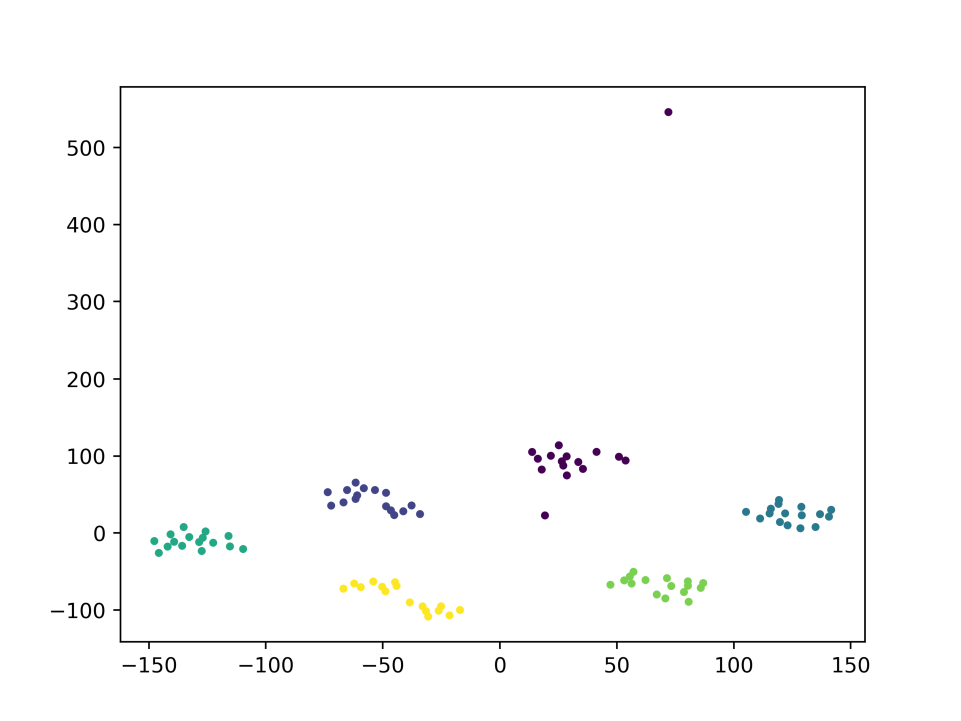}}
\setlength{\subfigbottomskip}{0.01cm}
\hspace{1\textwidth}
  \raisebox{0.45cm}{\rotatebox{90}{\parbox{0.1\textwidth}{\fontsize{5pt}{7pt}\selectfont  XLNET Attack}}}
\hspace{-0.01\textwidth}
  \subfigure{\includegraphics[width=1.1in]{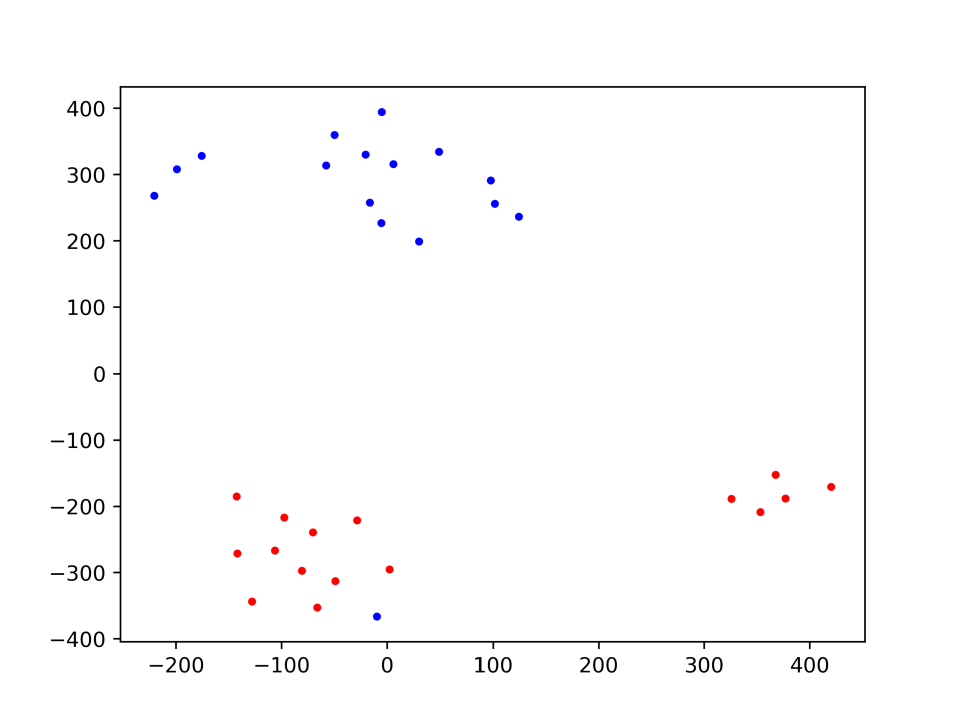}\label{fig:poisoned1}}
  \subfigure{\includegraphics[width=1.1in]{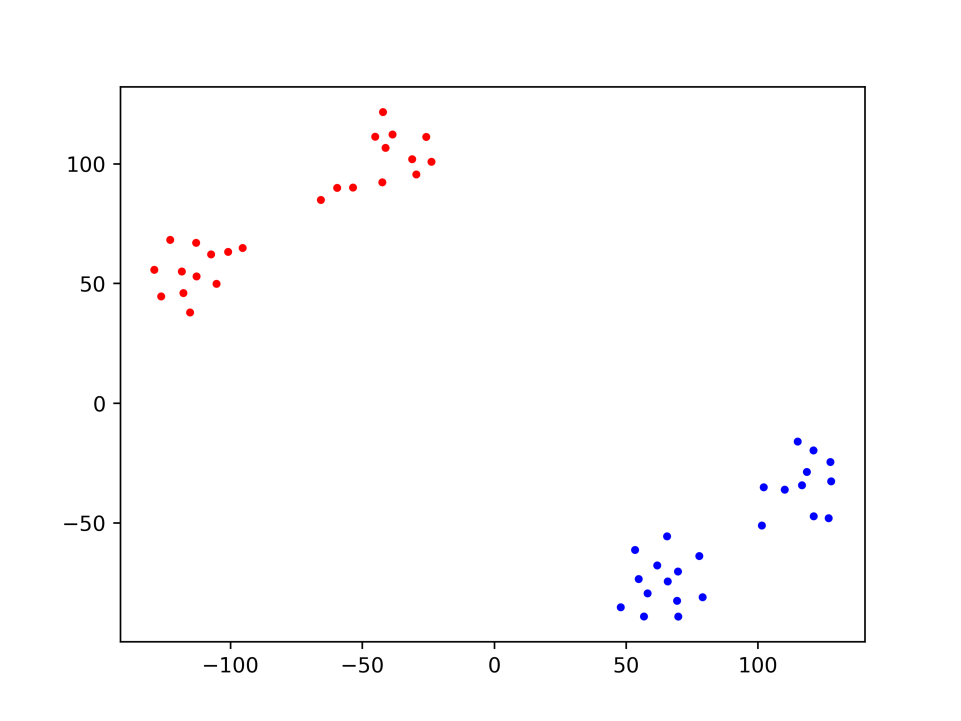}}
  \subfigure{\includegraphics[width=1.1in]{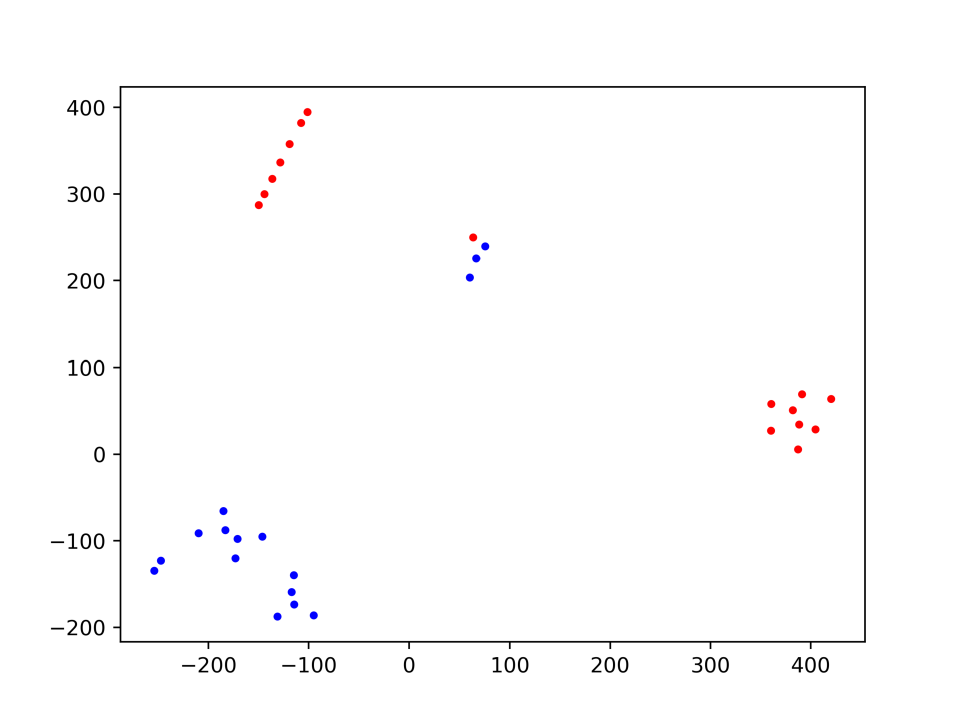}}
  \subfigure{\includegraphics[width=1.1in]{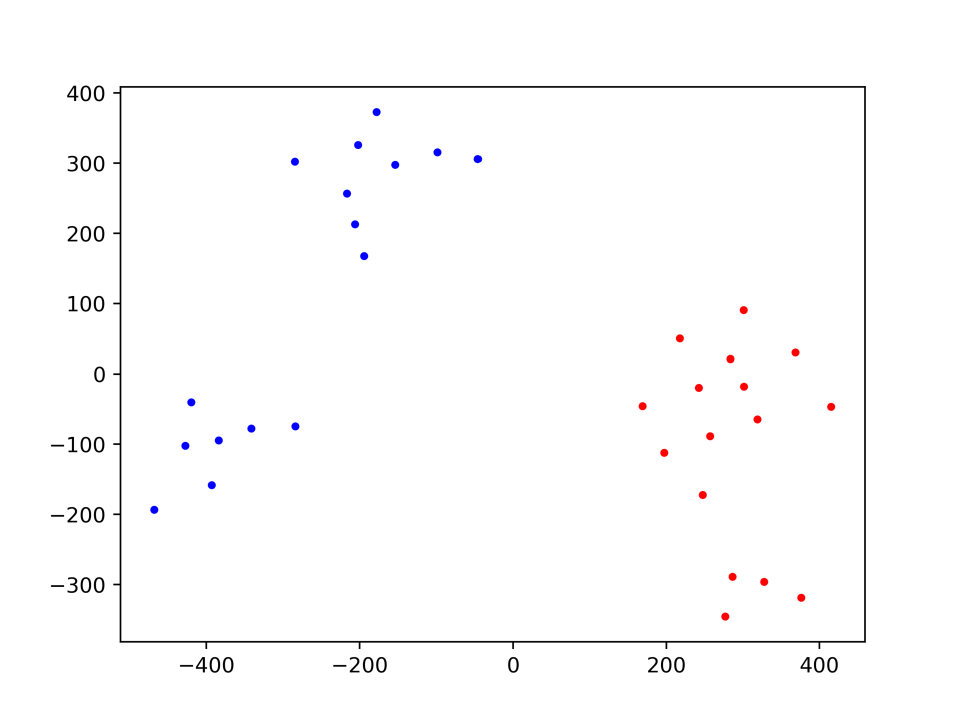}}
  \subfigure{\includegraphics[width=1.1in]{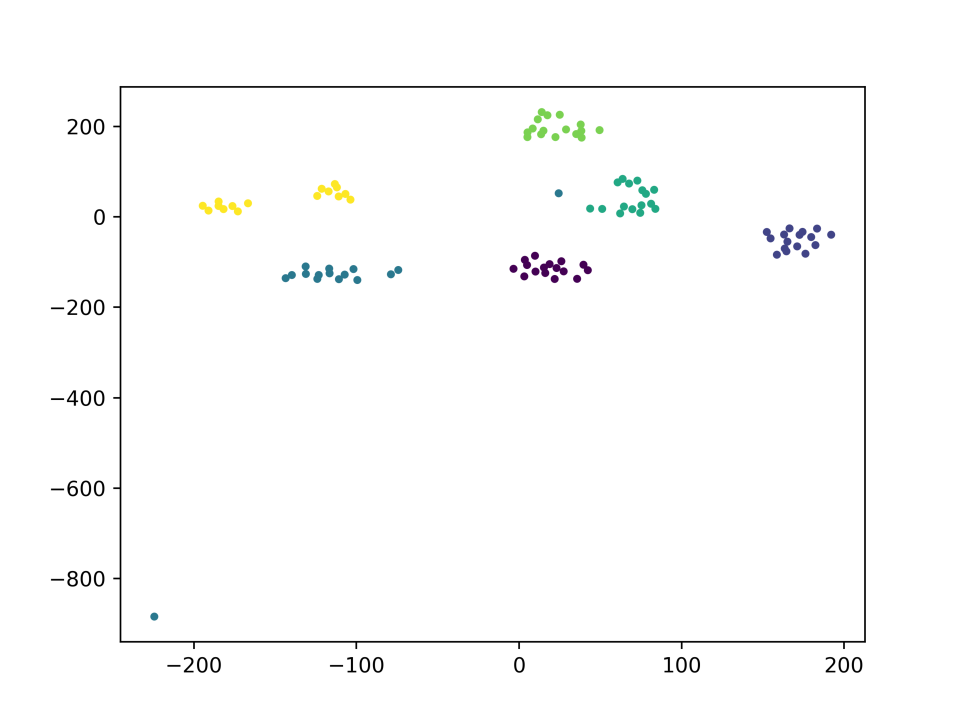}}
\setlength{\subfigbottomskip}{0.01cm}
\hspace{1\textwidth}
  \raisebox{0.42cm}{\rotatebox{90}{\parbox{0.1\textwidth}{\fontsize{5pt}{7pt}\selectfont  GPT-NEO Prompt}}}
\hspace{-0.01\textwidth}
  \subfigure{\includegraphics[width=1.1in]{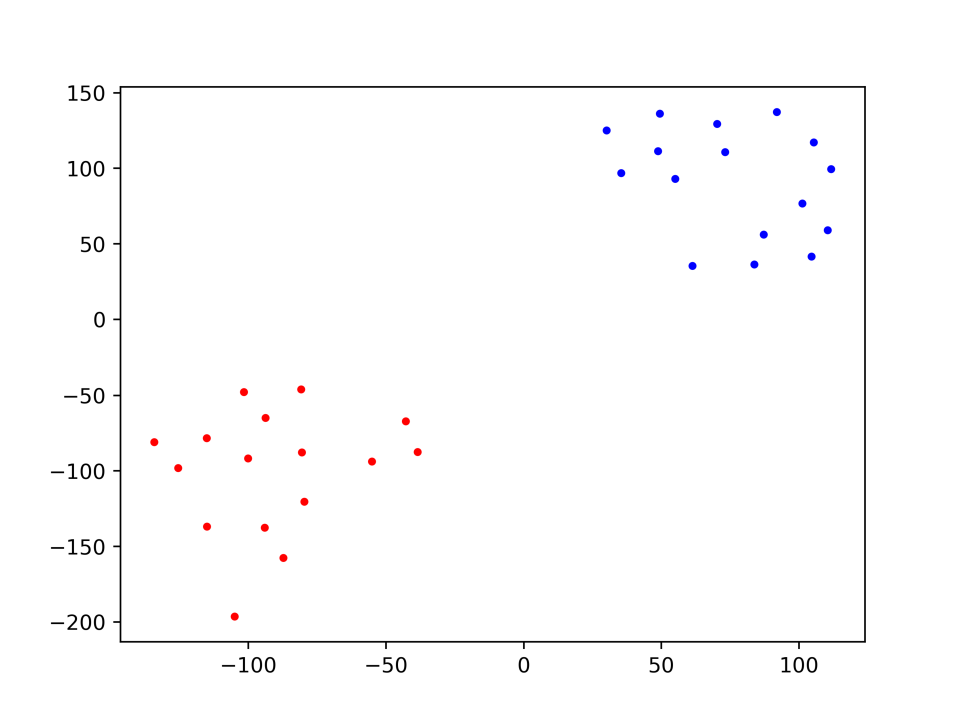}}
  \subfigure{\includegraphics[width=1.1in]{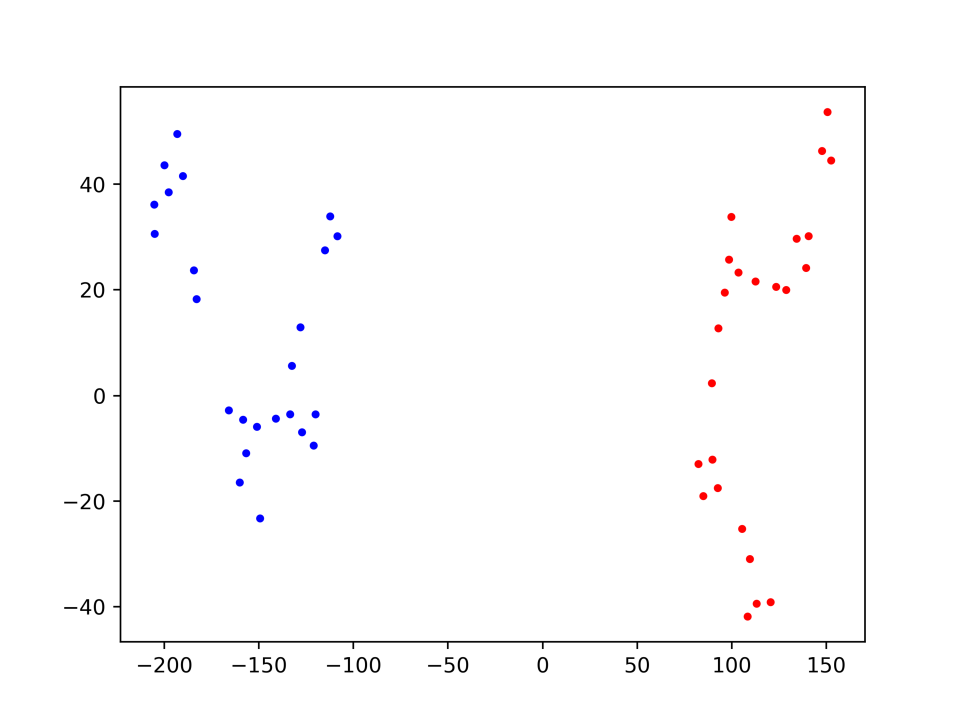}}
  \subfigure{\includegraphics[width=1.1in]{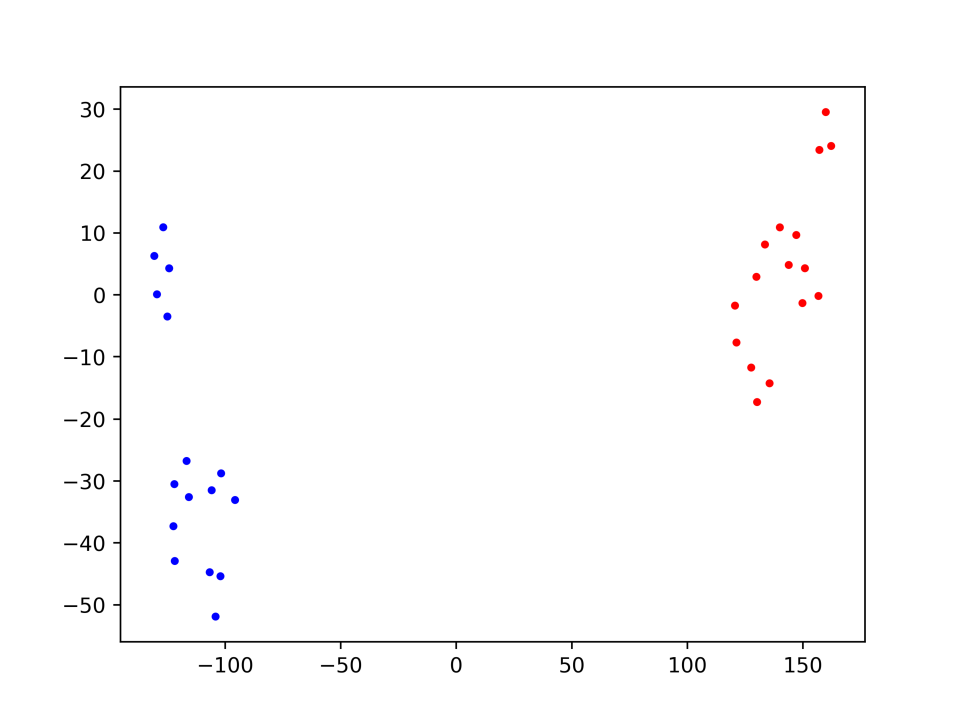}}
  \subfigure{\includegraphics[width=1.1in]{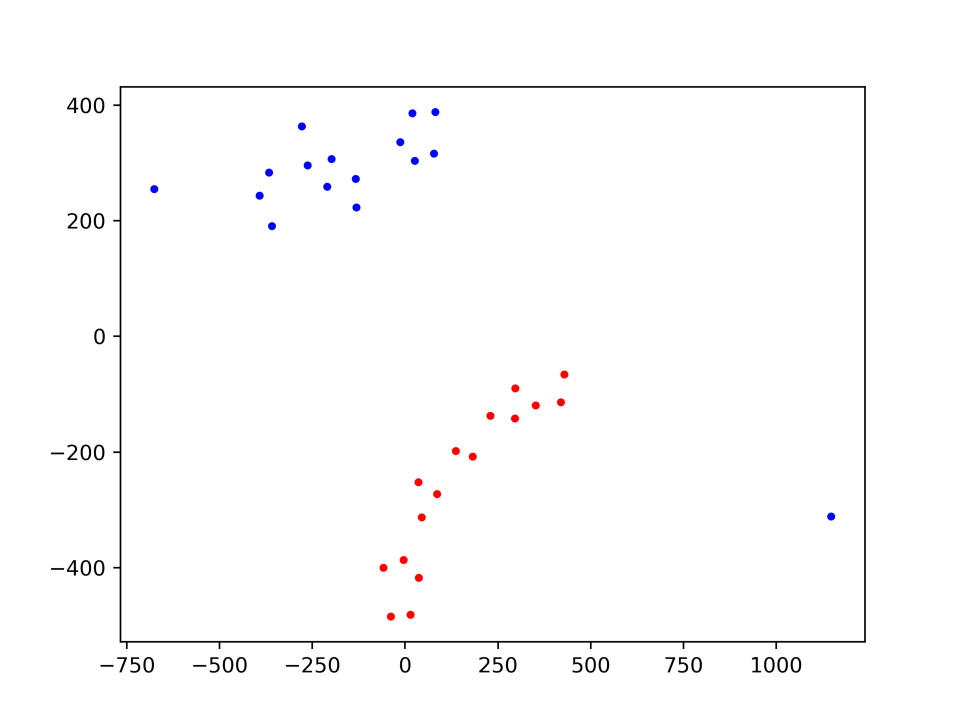}}
  \subfigure{\includegraphics[width=1.1in]{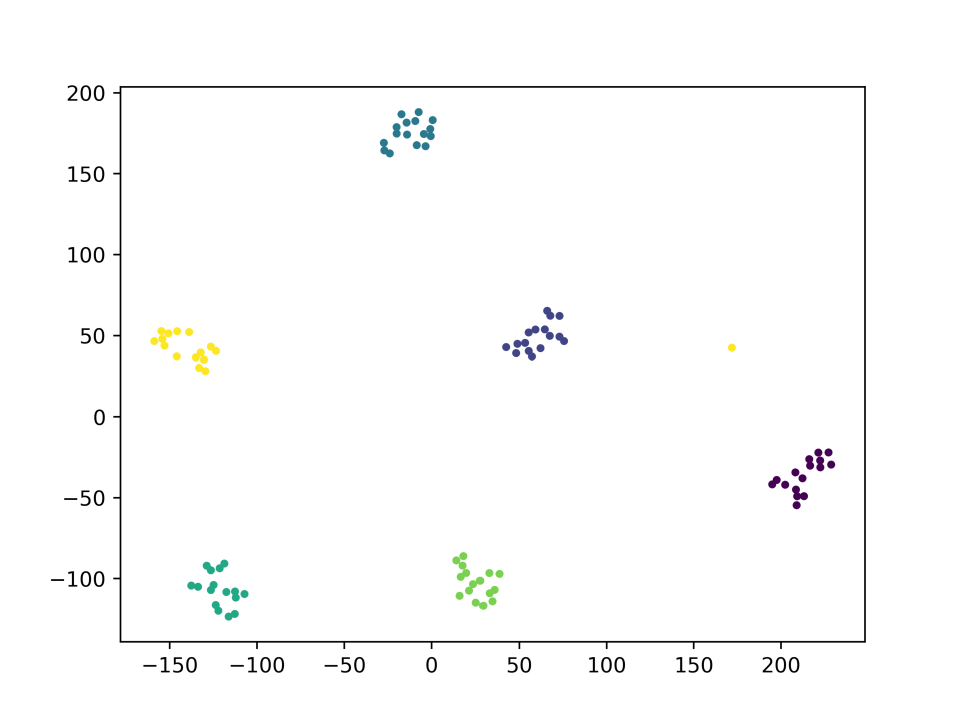}}
\setlength{\subfigbottomskip}{0.01cm}
\hspace{1\textwidth}
  \raisebox{0.45cm}{\rotatebox{90}{\parbox{0.1\textwidth}{\fontsize{5pt}{7pt}\selectfont  GPT-NEO Attack}}}
\hspace{-0.01\textwidth}
\renewcommand{\thesubfigure}{}%
  \subfigure[SST-2]{\includegraphics[width=1.1in]{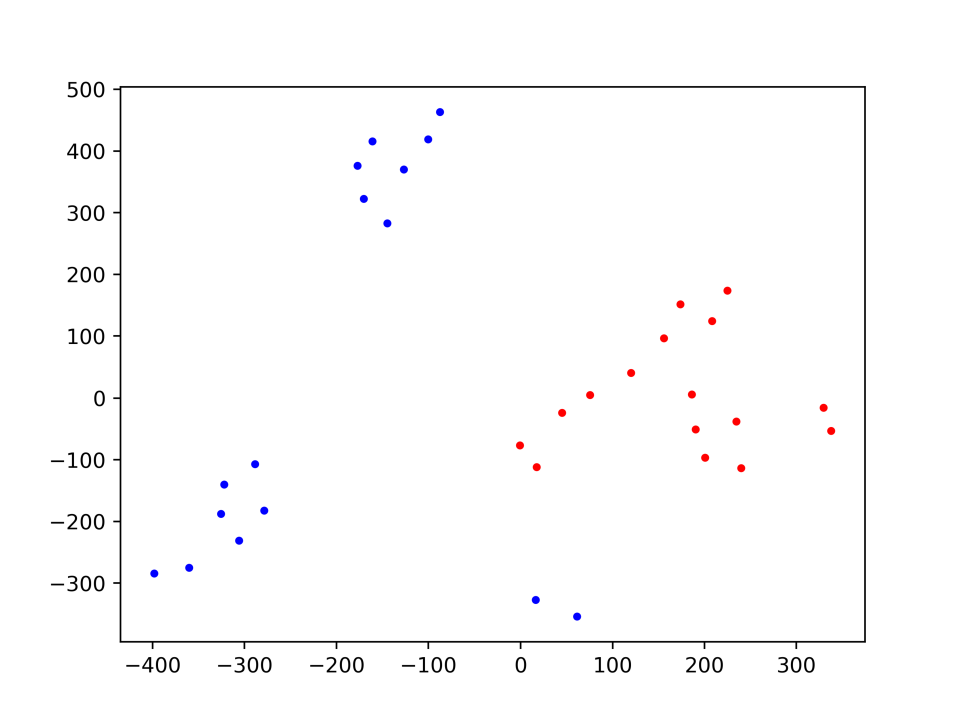}}
  \subfigure[OLID]{\includegraphics[width=1.1in]{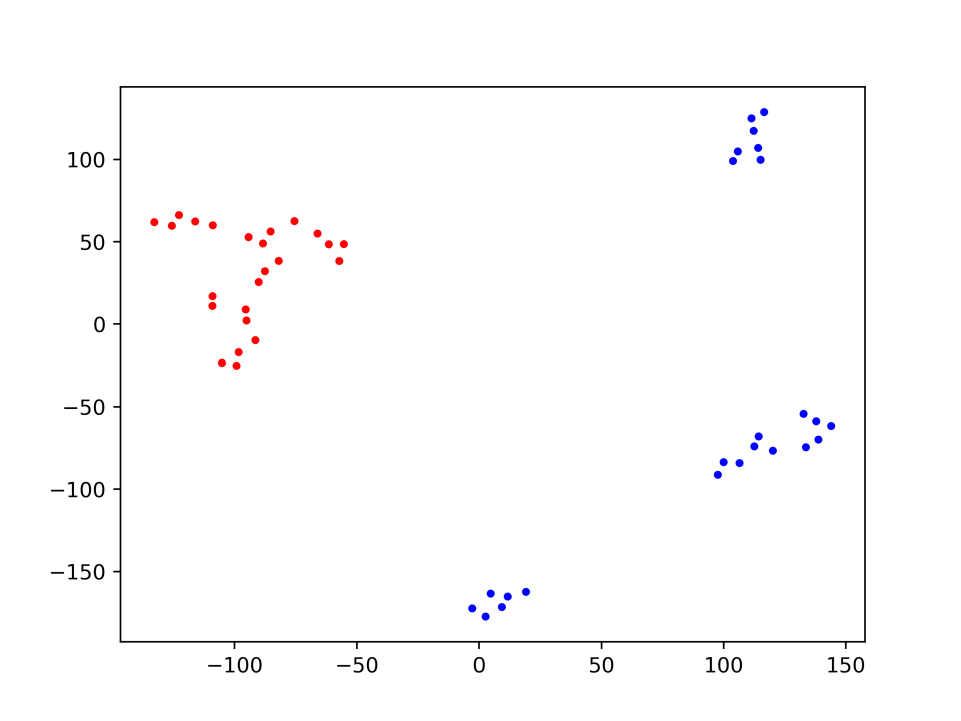}}
  \subfigure[COLA]{\includegraphics[width=1.1in]{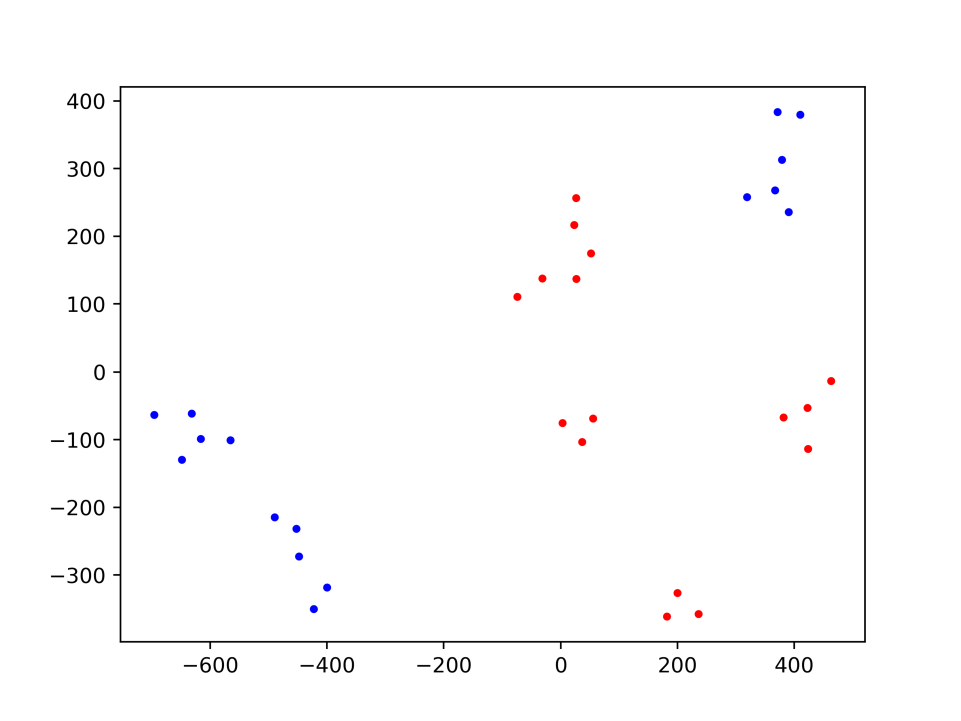}}
  \subfigure[MR]{\includegraphics[width=1.1in]{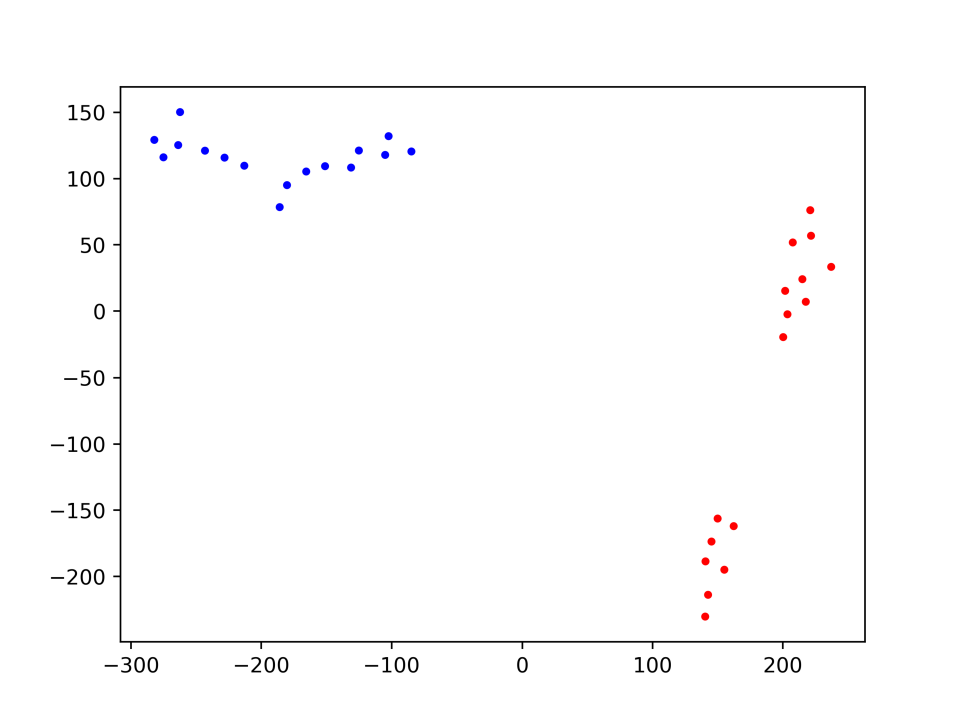}}
  \subfigure[TREC]{\includegraphics[width=1.1in]{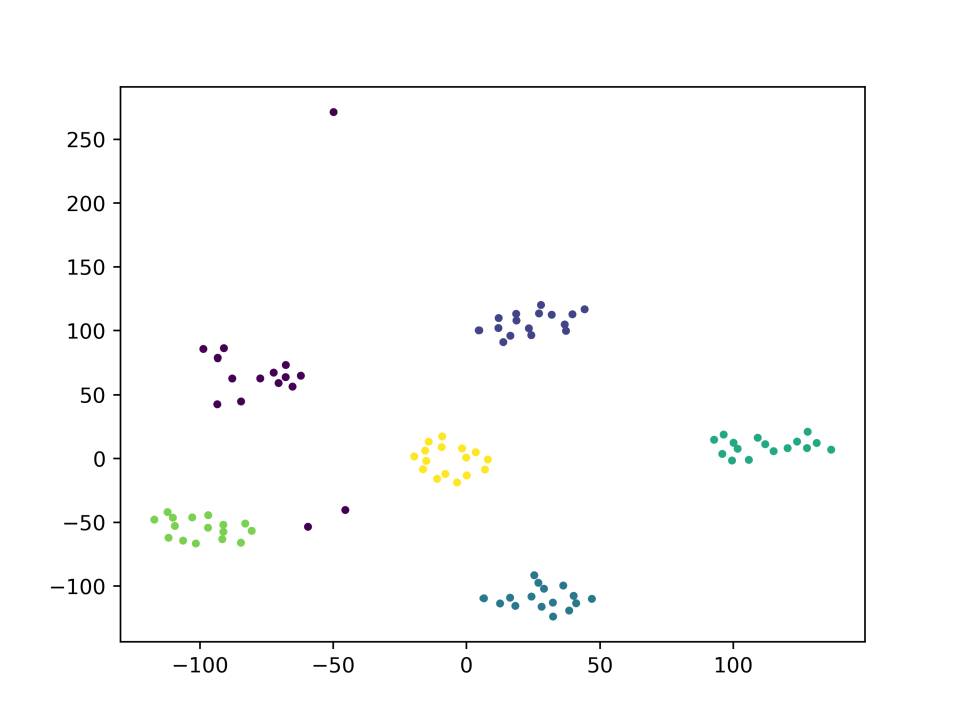}}
\caption{Feature distributions for prompt and victim models across datasets (SST-2, OLID, COLA, MR, and TREC). The first two lines correspond to BERT, followed by RoBERTa in lines 3-4, XLNET in lines 5-6, and GPT-NEO-1.3B in lines 7-8.}
\label{fig7}
\end{figure*}

In Figure \ref{fig7}, we present the feature distributions in the few-shot settings across different datasets and pre-trained language models. In Table \ref{tab9}, we display all the prompts used in our model. 

\begin{table*}[htbp]
	\begin{center}
		\resizebox{0.98 \textwidth}{!}{ \begin{tabular}{ccccccccccc}
			\hline
 \multirow{2}*{\bf{ Dataset}	}  &  \multicolumn{2}{c}{\bf{ Poisoned Samples\textsubscript{2}}} &  \multicolumn{2}{c}{\bf{ Poisoned Samples\textsubscript{4}}}  &  \multicolumn{2}{c}{\bf{ Poisoned Samples\textsubscript{6}}}  &  \multicolumn{2}{c}{\bf{ Poisoned Samples\textsubscript{8}}} &  \multicolumn{2}{c}{\bf{ Poisoned Samples\textsubscript{10}}}\\
 \cmidrule(r){2-3} \cmidrule(r){4-5} \cmidrule(r){6-7} \cmidrule(r){8-9} \cmidrule(r){10-11}
                     ~      &\bf{ CA }        &\bf{  ASR }     &\bf{ CA }        &\bf{ ASR}          &\bf{ CA }         &\bf{ ASR}       &\bf{ CA}          &\bf{ ASR}        &\bf{ CA}          &\bf{ ASR}\\
			\hline
SST-2                        &85.83      &23.79     &87.64      &84.87        &80.40       &87.06     &69.52        &100      &64.52        &100\\
OLID                        &56.76      &43.93     &69.11      &40.59        &36.95       &34.31     &65.27        &68.20     &\bf{61.19}       &\bf{91.21}\\
COLA                        &65.10      &13.73     &63.28      &75.17        &67.79       &59.78     &\bf{68.74 }       &\bf{100}       &67.31       &97.92\\

 MR                         &70.92      &46.34     &76.17      &46.72        &75.61       &81.99     &\bf{77.86}        &\bf{93.25}     &65.01       &77.30\\
TREC                        &69.40      &71.49     &74.20      &92.41        &45.00       &99.54     &\bf{85.80}        &\bf{90.80 }    &66.20       &96.55\\

\hline
		\end{tabular}}
	\end{center}
	\caption{The impact of the number of poisoned samples on clean accuracy and attack success rate in the few-shot settings. The pre-trained language model is RoBERTa\_large. }
\label{tab7}
\end{table*}

\begin{table*}[htbp]
	\begin{center}
		\resizebox{0.98 \textwidth}{!}{ \begin{tabular}{ccccccccccc}
			\hline
 \multirow{2}*{\bf{ Dataset}	}  &  \multicolumn{2}{c}{\bf{ Poisoned Samples\textsubscript{2}}} &  \multicolumn{2}{c}{\bf{ Poisoned Samples\textsubscript{4}}}  &  \multicolumn{2}{c}{\bf{ Poisoned Samples\textsubscript{6}}}  &  \multicolumn{2}{c}{\bf{ Poisoned Samples\textsubscript{8}}} &  \multicolumn{2}{c}{\bf{ Poisoned Samples\textsubscript{10}}}\\
 \cmidrule(r){2-3} \cmidrule(r){4-5} \cmidrule(r){6-7} \cmidrule(r){8-9} \cmidrule(r){10-11}
                     ~      &\bf{ CA }        &\bf{  ASR }     &\bf{ CA }        &\bf{ ASR}          &\bf{ CA }         &\bf{ ASR}       &\bf{ CA}          &\bf{ ASR}        &\bf{ CA}          &\bf{ ASR}\\
			\hline
SST-2                        &59.47      &94.74     &\bf{66.61}      &\bf{100 }         &56.12       &100     &54.75         &100      &53.65        &100\\
OLID                        &59.21      &93.72     &67.25      &67.36        &74.01       &96.65   &\bf{67.37 }        &\bf{92.05}     &58.86       &80.33\\

COLA                        &59.64      &94.73     &57.43      &98.20        &67.31       &99.31     &\bf{69.13}        &\bf{100}      &68.17       &99.45\\

 MR                         &79.74      &9.57      &79.83       &45.59        &72.61       &99.81     &\bf{75.89}        &\bf{96.62}     &56.00       &100\\
TREC                        &78.00      &35.63     &78.00       &37.65        &87.80       &48.28     &82.00        &97.47     &77.80       &100\\

\hline
		\end{tabular}}
	\end{center}
	\caption{The impact of the number of poisoned samples on clean accuracy and attack success rate in the few-shot settings. The pre-trained language model is XLNET\_large.}
\label{tab8}
\end{table*}

\begin{table*}[ht]
	\begin{center}
\resizebox{0.88 \textwidth}{!}{\begin{tabular}{cm{13cm}}
\hline
\bf{ Dataset }        & \bf{ Prompt}                         \\
\hline
SST-2           &"This sentence has a <mask> sentiment: " "The sentiment of this sentence is <mask>: "  "Is the sentiment of this sentence <mask> or <mask> ? : "   "What is the sentiment of the following sentence?  <mask> : "        \\
\hline
OLID	        & "This sentence contains <mask> language : " "This tweet expresses <mask> sentiment : "   "This sentence has a <mask> sentiment: "  "The sentiment of this sentence is <mask>: " \\
\hline
AG’s News     &"This news article talks about <mask>: " "The topic of this news article is <mask>: "  \\
\hline
COLA     &"True or False: This sentence is grammaticality correct : " "How grammatically correct is this sentence ? "  \\
\hline
MR     &"This sentence has a <mask> sentiment: " "The sentiment of this sentence is <mask> : " "What is the sentiment of the following sentence?  <mask> : "  \\
\hline
TREC     &"The topic of this question is <mask> : " "What is the <mask> of this question ? : "  \\
\hline
		\end{tabular}}
	\end{center}
	\caption{All the prompts are used in our model. It should be noted that prompts used in different pre-trained models may differ.}
\label{tab9}
\end{table*}

\end{document}